\documentclass[times,twocolumn,final]{elsarticle}
\usepackage{cag}
\usepackage{framed,multirow}
\usepackage{amssymb}
\usepackage{latexsym}
\usepackage{amsmath}
\usepackage{url}
\usepackage{xcolor}
\definecolor{newcolor}{rgb}{.8,.349,.1}

\usepackage{hyperref}

\newcommand{\State}{\mathcal S}
\newcommand{\Action}{\mathcal A}

\newcommand{\NN}{\mathbb N}
\newcommand{\RR}{\mathbb R}

\newcommand{\ie}{i.e.\ }
\newcommand{\eg}{e.g.\ }

\usepackage{subcaption}

\usepackage{enumitem}

\usepackage{soul}

\usepackage[switch,pagewise]{lineno} %Required by command \linenumbers below

\journal{Preprint}

\begin{document}

% \verso{Preprint Submitted for review}
% \verso{Understanding RL Crowds}
\verso{Kwiatkowski et al.}

\begin{frontmatter}

\title{\huge Understanding reinforcement learned crowds}%
% % \tnotetext[tnote1]{Only capitalize first
% % word and proper nouns in the title.}

\author[1]{Ariel Kwiatkowski\corref{cor1}}
\cortext[cor1]{Corresponding author.}
\emailauthor{ariel.kwiatkowski@polytechnique.edu}{A. Kwiatkowski}
%\ead{example@email.com}
    
\author[1]{Vicky Kalogeiton}
\author[2]{Julien Pettré}
\author[1]{Marie-Paule Cani}
% \fntext[fn1]{Footnote 1.}  

\address[1]{LIX, École Polytechnique/CNRS, Institut Polytechnique de Paris, Palaiseau, France}
\address[2]{INRIA Rennes, France}

%\received{1 February 2017}
% \received{\today}
%%%% Do not use the below for submitted manuscripts
%\finalform{28 March 2017}
%\accepted{2 April 2017}
%\availableonline{15 May 2017}
%\communicated{S. Sarkar}

\begin{abstract}
%%%
Simulating trajectories of virtual crowds is a commonly encountered task in Computer Graphics.
Several recent works have applied Reinforcement Learning methods to animate virtual agents, however they often make different design choices when it comes to the fundamental simulation setup. Each of these choices comes with a reasonable justification for its use, so it is not obvious what is their real impact, and how they affect the results.
In this work, we analyze some of these arbitrary choices in terms of their impact on the learning performance, as well as the quality of the resulting simulation measured in terms of the energy efficiency. 
We perform a theoretical analysis of the properties of the reward function design, and empirically evaluate the impact of using certain observation and action spaces on a variety of scenarios, with the reward function and energy usage as metrics. 
We show that directly using the neighboring agents' information as observation generally outperforms the more widely used raycasting. Similarly, using nonholonomic controls with egocentric observations tends to produce more efficient behaviors than holonomic controls with absolute observations.
Each of these choices has a significant, and potentially nontrivial impact on the results, and so researchers should be mindful about choosing and reporting them in their work.
%%%%
\end{abstract}

\begin{keyword}
%% Keywords
\KWD Reinforcement learning\sep Crowd simulation\sep Character animation\sep Multiagent reinforcement learning\sep Artificial intelligence
% Computers and Graphics\sep Formatting\sep Guidelines
\end{keyword}

\end{frontmatter}

% \linenumbers


\begin{figure*}
  \includegraphics[width=\textwidth]{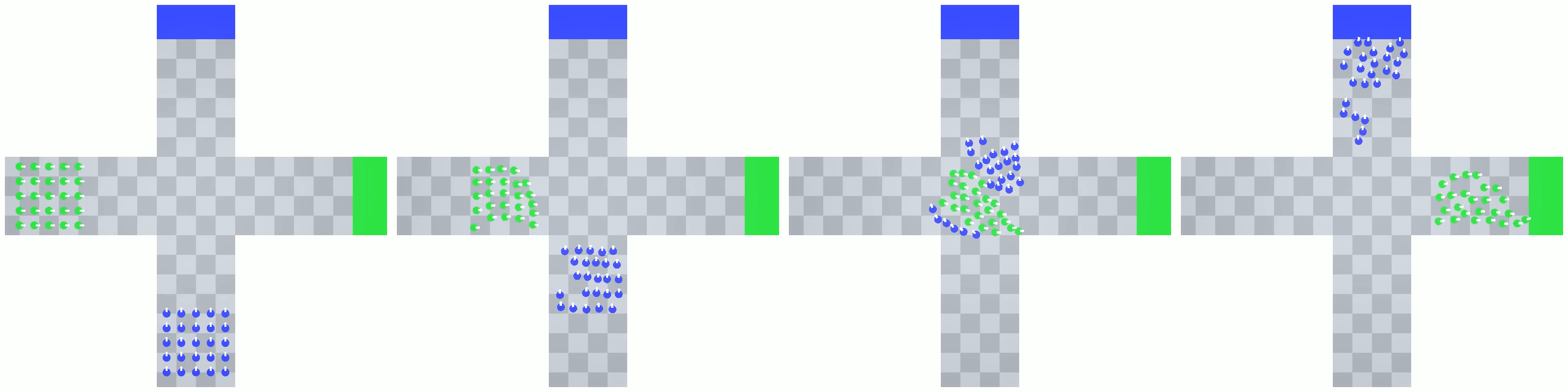}
  \caption{Two groups of virtual agents crossing a corridor}
  \label{fig:teaser}
\end{figure*}

\section{Introduction}

% Simulating crowds
Simulating virtual human crowds is a common task when creating lively, populated scenes for graphics applications. %There is a large diversity of scenarios that this includes
This includes a large variety of scenarios -- ranging from small, artificially structured scenes used in research, to large scale simulations with thousands of virtual agents. They can be used either for real-life applications (\eg testing evacuation scenarios in airports or sport stadiums) or targeting automatic content creation in films and games, featuring a diversity of background characters, each with their own goals and motivations.  %enormous real-life situations involving hundreds of people, each with their own goals and motivations, such as those used for testing evacuation scenarios in airports or sport stadiums. 

% Classic approaches -> RL approaches
While the approaches to simulating such crowds typically use hand-crafted or data-driven algorithms, in recent years there has been an increasing research interest in using Deep Reinforcement Learning (DRL) methods. These algorithms have a promise of a flexible, problem-agnostic training process that can autonomously produce diverse behaviors. However, they also bring many new challenges, which must be explored independently of the previous knowledge of crowd simulation algorithms.

% Design of RL env, obs, actions
It is worth noting that just ``crowd simulation'' is somewhat of an underdefined problem. Although we have an intuition as to what a crowd is and how it behaves, it is not clear how to formalize it, which is exemplified by the variety of descriptions in prior work.
%as summarized in Section~\ref{sec:background}. 
At the same time, as Reinforcement Learning (RL) is designed to optimize a given scalar reward function, it excels when the objective is clearly stated. For this reason, we investigate the various ways to specify the crowd simulation problem. 

There are three main components of an RL system -- observations, actions, and rewards. For each of them, it is not necessarily obvious what is the appropriate level of abstraction, or even the implementation, to properly simulate human crowds. Take actions, for example -- it is infeasible to perform a fully accurate biomechanical simulation of each muscle movement, so we must use a simplified model. Similarly for observations, performing a full rendering of the scene to emulate human vision would not only be expensive to perform, it would also difficult to train.

% Reward tuning
Designing the reward function is arguably even more complex. While navigating in our daily life, we balance several, often mutually contradictory objectives, such as getting to the destination efficiently, moving at a comfortable speed or avoiding bumping into others. Even though some of them are simple to formalize as a single scalar reward, it is not obvious how they should be balanced. Is it acceptable to reach the destination two seconds earlier, but increase the risk of bumping into someone by 5\%? Can we take a shorter path which leads through a group of people, or should we go around, reducing risk of collision, but increasing the distance? The answer to each of this question will vary from person to person, and many external factors. If going a bit slower would cause us to barely miss a train, we are likely to accept a higher risk of colliding than if we are just taking a walk around the park.

In this work, we intend to bring clarity to the intersection of crowd simulation and DRL, by exploring in detail the impact of these design choices on the generated virtual crowds. We perform a theoretical analysis of controlling the agents' velocities, and an empirical investigation of various observation and action spaces. We evaluate them in terms of optimizing the reward function, but also consider the energy expenditure, and various quantitative properties of the movement. Our contributions are:
\begin{enumerate}
    \item Empirical evaluation of raycasting versus direct agent perception in RL crowd simulation
    \item Empirical comparison of various implementations of observations and dynamics in RL crowd simulation
    \item Theoretical and empirical analysis of the properties of reward functions for efficient navigation with RL
\end{enumerate}

\section{Background \& Related Work}\label{sec:background}

Microscopic simulation of virtual crowds has garnered significant research interest in recent years. Various non-learning techniques have been introduced in the previous decade~\citep{toll_algorithms_2021}, which typically involve designing rule-based systems to update each agent's velocity based on their context. There also exists a body of work using RL for controlling virtual characters, including crowd scenarios~\citep{kwiatkowski_survey_2022}. In this section we describe the elements of related work which are the most relevant to applying RL in crowd simulation.

% \vspace{2mm}
\textbf{Reinforcement Learning.}
RL is a study of sequential decision making, where one or more agents act in an environment, affecting its state, and receiving rewards. Modern RL widely uses neural network-based algorithms~\citep{sutton_reinforcement_2018}, using them \eg as a policy function, mapping observations to actions taken by the agent. This network is then optimized with a procedure based on the Policy Gradient Theorem, as introduced by the REINFORCE algorithm~\citep{sutton_policy_1999}.
A more modern version that follows the same principle is the Proximal Policy Optimization (PPO) algorithm, introduced by \citet{schulman_proximal_2017}. PPO is now the de facto standard on-policy algorithm used in many DRL applications due to its simplicity and efficiency.

% \vspace{2mm}
\textbf{Design Choices in RL.}
The importance of various, seemingly minor design choices in RL is a common phenomenon across many domains of the field. In physics-based animation, \citet{reda_learning_2020} explore how the performance of RL agents is affected by parameters like the initial state distribution, components of the state representation, or the control frequency. \citet{engstrom_implementation_2020} perform a large-scale study of implementation details and code-level optimizations that affect the performance of TRPO and PPO in robotic control tasks. Similarly, \citet{andrychowicz_what_2020} investigate how different design choices in on-policy algorithms affect the agent's performance.

% \vspace{2mm}
\textbf{Crowd Simulation.} 
 Microscopic simulation of crowds is typically done by either using \textbf{force-based} methods~\citep{helbing_social_1995} where the positions of nearby agents, obstacles, as well as the agent's destination, all contribute to Social Forces that drive the acceleration of the agent at the next time step, or alternatively \textbf{velocity-based} methods such as Optimal Reciprocal Collision Avoidance (ORCA)~\citep{siciliano_reciprocal_2011}. The latter construct obstacles in the agent's velocity space, which correspond to velocities that would result in a collision if the agent were to take them. This leads to effective solutions to collision avoidance, able to capture  anticipation behaviours in crossing scenarios.
Recently, \textbf{vision-based} and \textbf{data-driven} algorithms were explored as well,
%are also becoming relevant, 
promising even more human-like results~\citep{toll_algorithms_2021}.

%%% Previous version from Ariel
%There are many approaches for microscopic simulation of crowds. The main two categories are \textbf{force-based} and \textbf{velocity-based} methods. Recently, \textbf{vision-based} and \textbf{data-driven} algorithms are also becoming relevant, promising more human-like results~\citep{toll_algorithms_2021}.

%In velocity-based methods, a commonly used algorithm for pedestrian simulation is Optimal Reciprocal Collision Avoidance (ORCA)~\citep{siciliano_reciprocal_2011}, which builds upon the Reciprocal Velocity Obstacle algorithm~\cite{van_den_berg_reciprocal_2008}. Both of these methods work by constructing obstacles in the velocity spaces, which correspond to velocities that would result in a collision if the agent were to take them. If all agents are controlled by the same method, then the reciprocity approach provides a guarantee of having a collision-free trajectory if such trajectory exists. 

%Force-based methods update the agent's velocity indirectly. In the Social Force model~\citep{helbing_social_1995}, the positions of other nearby agents, obstacles, as well as the agent's destination, all induce an acceleration on the agent. Then it follows the velocity which results from integrating that acceleration over time. This simple model makes it possible to create diverse agents with varied motivations, and smooth, collision-free trajectories.

% \vspace{2mm}
\textbf{Crowd Simulation via DRL.} 
There is a number of prior papers which train DRL agents on the task of crowd simulation. \citet{long_towards_2018} use a multiagent robotic navigation setup, which shares certain properties with crowd simulation. \citet{lee_crowd_2018} train 
%a DDPG~\citep{lillicrap_continuous_2015} agent 
an RL agent
on a variety of crowd scenarios, showing that a single trained model can be used to control multiple agents acting in a shared environment, on a diverse range of scenarios. \citet{sun_crowd_2019} train groups of agents by making them follow specially-trained leader agents. \citet{xu_local_2020} combine DRL with an ORCA layer that ensures collision-free movement.

DRL is also used to generate more interesting, higher-quality trajectories. \citet{xu_human-inspired_2021} use real-world human trajectory data to train a supervised model judging the human-likeness of a generated trajectory. Then, the output of that model is used as an additional component of the reward function, encouraging agents to act in a human-like manner. \citet{hu_heterogeneous_2022} use a parametric RL approach to generate heterogeneous behaviors with a single shared policy network. Each agent has its own preferred velocity value, and is trained to move according to it. Similarly, \citet{panayiotou_ccp_2022} vary the weights in the reward functions of different agents in order to give them unique and configurable personalities.

While each of these prior works tackles the same problem of crowd simulation, many elements of their fundamental setup differ in potentially significant ways, which makes direct comparison infeasible. Therefore, in this work, we compare the basic design choices used in all these papers, in order to analyse their impact.
%In contrast, in this work, we compare the basic design choices used in all these papers, in order to analyse their impact. In Section~\ref{sec:choices}, we detail the specific choices used in prior work that we explore in this paper.

\section{Environment Design Choices}\label{sec:choices}

We identify three design choices which can impact the properties of virtual crowds trained with a standard DRL algorithm -- observations, actions, and the reward function. In this section, we set the problem in standard multiagent RL formalism, and describe the variants of observations and action spaces explored in this work. 
%We look at each of these aspects in detail, and describe their respective variants.

\subsection{Problem Formulation}

We model the problem of crowd simulation as a Partially Observable Stochastic Game (POSG)~\cite{hansen_dynamic_2004}. A POSG is defined as a tuple $(\mathcal{I}, \State, \{ \Action_i \}, \{\Omega_i\}, \{O_i\}, T, \{R_i\}, \mu)$, where $\mathcal{I}$ is the set of agents, $\State$ is a set of states of the environment, $\Action^i$ is a set of actions for agent $i$ ($\Action = \times_{i \in \mathcal{I}} \Action^i$ is the joint action set), $\Omega^i$ is the set of observations, $O^i\colon \State \to \Omega_i$ is the observation function, $O^i\colon \State \to \Omega_i$ is the observation function, $T\colon \State \times \Action \to \Delta \State$ is the environment transition function, $R^i\colon \State \times \Action \times \State \to \RR$ is the reward function, and $\mu \in \Delta \State$ is the initial state distribution.

% \begin{itemize}
%     \item $\mathcal{I}$ is the finite set of agents, indexed $1, \dots, n$
%     \item $\State$ is a set of states of the shared environment.
%     \item $\Action^i$ is a set of actions available to agent $i$, and $\Action = \times_{i \in \mathcal{I}} \Action^i$ is the joint action set.
%     \item $\Omega^i$ is the set of observations available to agent $i$.
%     \item $O^i\colon \State \to \Omega_i$ is the observation function for agent $i$.
%     \item $T\colon \State \times \Action \to \Delta \State$ is the environment transition function, representing its dynamics.
%     \item $R^i\colon \State \times \Action \times \State \to \RR$ is the reward function of agent $i$, which defines the agent's task.
%     \item $\mu \in \Delta \State$ is the initial state distribution.
% \end{itemize}

In a POSG, all agents simultaneously make decisions based on their own private observations. Then, the environment is updated according to the joint action of all agents, and each agent receives its own reward that it tries to maximize. 
The reward is computed the same way for each agent, but based on its individual situation (\ie no reward sharing). 
We additionally specify a time limit $T_{max} \in \NN$ which is the maximum number of steps the environment is allowed to take before resetting.

\subsection{Observation Space}

In order to navigate through the environment, each agent must perceive its environment in some way. However, it is not obvious in what form agents should receive this information, or in fact, what the information should be. 

The simplest human-inspired design is to give the agent information in its own frame of reference, and to have it perceive the environment through raycasting -- a simple approximation of human vision. The intention is that if humans can effectively navigate using this type of information, then it should also suffice for virtual agents, which should then act more human-like. However, it is not necessarily the case that an anthropomorphic structure is indeed optimal for virtual agents, especially with it being only a rough approximation. More realistic rendering of the agent's vision is an option, but it would result in very large observation sizes, subject to the curse of dimensionality. Thus, it is worthwhile to explore other possibilities.

An agent must receive information about its surroundings (other nearby agents and obstacles, \ie \textbf{Environment Perception}), but also about its own internal state and knowledge (\eg its current velocity, or its current destination, \ie \textbf{Proprioception}). In both of these cases, it is also relevant what is the \textbf{reference frame} in which they are observed.
% For the former, we consider three representations: absolute, relative, and egocentric.
For the Environment Perception, we consider two types of perception: Raycasting and Direct Agent Perception (AP) (Section~\ref{sec:env-perception}). For the reference frame, we have three representations: Absolute, Relative, and Egocentric (Section~\ref{sec:proprioception}).

\subsubsection{Environment Perception}\label{sec:env-perception}

Raycasting refers to a method where several rays are cast from the center of the agent, in a plane parallel to the ground. Each of those rays then provides information on whether or not it collides with any object within a predefined distance, and what is the distance to the collision location. 

% TODO: probably split/rephrase this?
Direct Agent Perception, similar to the method used by \citet{xu_human-inspired_2021} is an alternative approach, where the agent directly receives the positions of other agents within a certain range, along with other relevant parameters. 
The reference frame in which the information is passed follows the chosen proprioception model (\ie Absolute, Relative or Egocentric).
This has the possible benefit of directly giving access to relevant information, but it introduces two important complications. Firstly, this method cannot canonically represent obstacles. While small, human-sized obstacles can be treated as stationary agents, large obstacles like walls need a different approach. Secondly, the number of neighboring agents is variable, and can grow very large in high density scenarios. The standard Multi-Layer Perceptron architecture cannot handle variable-sized inputs, so this variability has to be accounted for in some way. Furthermore, the order in which neighboring agents are observed is irrelevant, so a permutation-invariant architecture is necessary to decrease the effective size of the observation space.

In this work, we also test a multimodal approach which combines both Raycasting and Agent Perception, described in more detail in Section~\ref{sec:architecture}. With this hybrid method, the raycasting is only used to perceive static obstacles such as walls, ignoring other agents, whose positions are instead observed directly. We hypothesize that this might enable gaining the benefits of both methods -- the agent has accurate knowledge of others, as well as a general idea of the surrounding layout, sufficient for navigation.

\subsubsection{Reference Frames}\label{sec:proprioception}

Absolute observations~\citep{sun_crowd_2019} use a bird's-eye view on the global scene. The agent observes a vector consisting of its position $\mathbf p$ and the position of its goal $\mathbf p_g$ in the global coordinate frame, its current orientation $\phi$ (which is relevant for some choices in Action Spaces), and its current velocity $\mathbf v$. Using the Agent Perception approach, the agent observes the positions $\mathbf p_i$ and velocities $\mathbf v_i$ of nearby agents in the global frame.

Relative observations~\citep{xu_human-inspired_2021, long_towards_2018, xu_local_2020}, like Absolute, use the global frame, however it is translated so that it is centered on the agent. It again receives its absolute position $\mathbf p$ and velocity $\mathbf v$ in order to retain the information about the large context, but the goal position is given relatively to the agent's own position, as $\mathbf p_g - \mathbf p$. Similarly, using Agent Perception, other agents' positions are given as $\mathbf p_i - \mathbf p$, and velocities are given in the absolute form $\mathbf v_i$.

Egocentric observations~\citep{hu_heterogeneous_2022, lee_crowd_2018} use the agent's local frame according to its orientation. The agent observes its position $\mathbf p$ and orientation angle $\phi$ in the global frame. We write $R_\phi$ to be the rotation matrix associated with the agent's current orientation. The agent observes its goal position as $R_\phi (\mathbf p_g - \mathbf p)$. Using Agent Perception, the positions of other agents are also represented as $R_\phi (\mathbf p_i - \mathbf p)$, and their observed velocities are $R_\phi \mathbf v$.

While each of these has a theoretical justification (moving from absolute information about the scene, towards a more human-like first-person view), it is not immediately obvious which one is the best. On the one hand, absolute observations can give a high-level overall view of the scene, potentially aiding coordination. On the other hand, a relative or egocentric view allows agents to better reuse experiences between different positions and situations. If the agent is heading towards its goal, it is not that important whether it is to the left or to the right, looking from a bird's eye view. Since the Egocentric observations lose this information, navigation might be expected to be learned more efficiently. Note that the choice of the reference frame also affects the structure of Agent Perception.

\subsection{Action Space and Dynamics}

\begin{figure}
\centering
\begin{subfigure}{0.11\textwidth}
    \includegraphics[width=\textwidth]{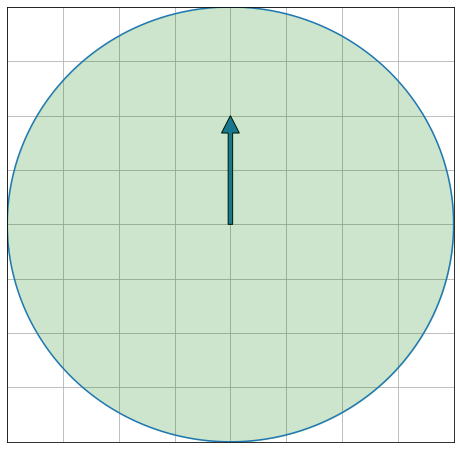}
    \caption{Cartesian Velocity.}
    \label{fig:carvel}
\end{subfigure}
\hfill
\begin{subfigure}{0.11\textwidth}
    \includegraphics[width=\textwidth]{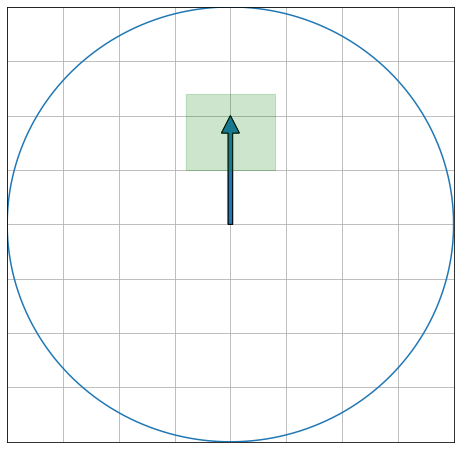}
    \caption{Cartesian Acceleration.}
    \label{fig:caracc}
\end{subfigure}
\hfill
\begin{subfigure}{0.11\textwidth}
    \includegraphics[width=\textwidth]{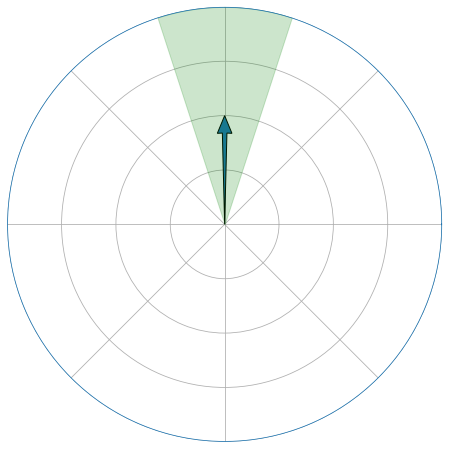}
    \caption{Polar Velocity.}
    \label{fig:polvel}
\end{subfigure}
\hfill
\begin{subfigure}{0.11\textwidth}
    \includegraphics[width=\textwidth]{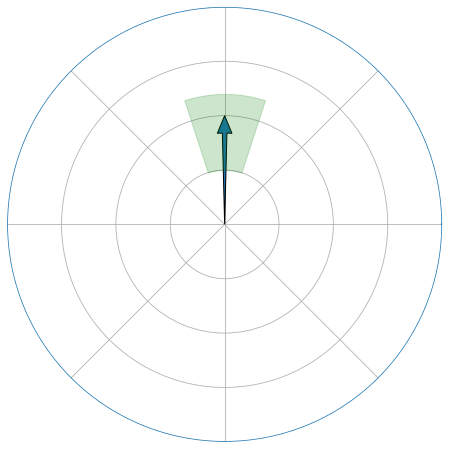}
    \caption{Polar Acceleration.}
    \label{fig:polacc}
\end{subfigure}
        
\caption{A schematic representation of the available action spaces. In each case, we take a bird's-eye view of an agent moving in the positive Y direction at an intermediate speed, represented by the blue arrow. The blue circle represents the space of all physically possible velocities (\ie below the maximum speed). The green area represents the velocities that the agent is able to have in the following timestep under the specific action space.}
\label{fig:action-spaces}
\end{figure}

Human motion is highly complex, and a biomechanically accurate simulation of human motion is a challenging research problem in of itself, so for the purposes of creating virtual crowds, we use a simplified model. The simplest choice is holonomic locomotion~\citep{hughes_holonomic_2015}, where at each step, the agent can choose its velocity constrained only by its magnitude. However, this approach does not correspond well to the motion constraints of real humans. \citet{arechavaleta_nonholonomic_2008} propose a nonholonomic model, in which an agent can move in the direction of its current orientation, and incrementally change its orientation for the next timestep. 

% Do I cite newtonian mechanics?
Allowing the agent to freely choose its velocity at every timestep gives it much more flexibility in choosing its behavior. However, from the perspective of Newtonian mechanics, it is more physically justified for the agent to directly choose its acceleration. This would mean that the velocity change at each timestep is incremental.

Similarly, there is a choice in how the actions should be represented. We can take the bird's eye view, where the agents choose their actions according to an absolute reference frame, moving up or down, left or right. Alternatively, we can take a more individual perspective, where the agents operate in a polar frame, choosing their linear movement and the direction of that movement.

For this reason, we consider four different dynamics models of the environment: Cartesian Velocity, Cartesian Acceleration, Polar Velocity, and Polar Acceleration.

Cartesian control (used by \citep{xu_human-inspired_2021, hu_heterogeneous_2022}) implies that the agent separately chooses the $x$ and $y$ components of its motion - either of its velocity or acceleration.

Polar control (used by \citep{lee_crowd_2018, long_towards_2018, sun_crowd_2019, xu_local_2020}), implies that the agent separately updates its orientation angle, and its linear speed. The linear speed is again updated either by choosing an arbitrary value below a certain magnitude (velocity controls), or by incrementally updating it according to the acceleration chosen by the agent.

Acceleration controls 
(used by \citep{hu_heterogeneous_2022, xu_human-inspired_2021}),
%% MP: this inof is already given above
are modeled in our implementation using a linear damping model.
This means that given an acceleration $a_t$ chosen by the agent as its action, the updated velocity will be 
% $v_t = v_{t-1} + (a_t - \lambda v_{t-1}) \Delta t$
\[
    v_t = v_{t-1} + (a_t - \lambda v_{t-1}) \Delta t
\]
% Should I elaborate on how \lambda is computed? It's a pretty simple physical calculation, we're basically looking for a_max = \lambda v_max
where $\lambda$ is chosen so that
%, given a maximum acceleration value, 
we obtain the same maximum speed as in velocity controls, which is equal to $2 \frac{m}{s}$. 

With velocity controls 
(used by \citep{lee_crowd_2018, long_towards_2018, sun_crowd_2019, xu_local_2020}), the agent can choose an arbitrary speed of a magnitude lower than $2 \frac{m}{s}$.

In Cartesian controls, the agent's orientation is defined to be parallel to its current velocity. In Polar controls, the orientation is directly controlled, and the velocity is parallel to the orientation.

\section{Reward Function Design}\label{sec:reward}

Designing the reward function is arguably the most impactful, and the most difficult part of creating an RL-driven crowd simulation. The two components which are present in all existing work are a positive term correlated with reaching the goal, and a negative term correlated with collisions. Beyond that, various elements may be included to promote certain behavior characteristics, or to improve training performance through reward shaping. In this work, we consider the following reward components based on prior work:

\begin{enumerate}
    \item Reward for reaching the goal $R_g = c_g$ (once) \citep{long_towards_2018, sun_crowd_2019, xu_local_2020}
    \item Reward for approaching the goal $R_p = c_p (d_t - d_{t-1})$ (every timestep) \citep{xu_human-inspired_2021, hu_heterogeneous_2022, lee_crowd_2018, long_towards_2018, sun_crowd_2019, xu_local_2020}
    \item Reward for maintaining a comfortable speed $R_v = -c_v |v - v_0|^{c_e}$ (every timestep) \citep{xu_human-inspired_2021}
    \item Penalty for collisions $R_c = -c_c$ (every collision, every timestep) \citep{xu_human-inspired_2021, hu_heterogeneous_2022, lee_crowd_2018, long_towards_2018, sun_crowd_2019, xu_local_2020}
    \item Reward for urgency $R_t = -c_t$ (every timestep) \citep{sun_crowd_2019}
\end{enumerate}
where $c_g$, $c_p$, $c_v$, $c_e$, $c_c$, $c_t$ are arbitrary (typically positive) coefficients. Their roles are as follows: $R_g$ is the main (sparse) reward representing the agent's destination. $R_p$ provides a dense reward for the navigation objective to enable faster training. $R_v$ incentivizes moving at a comfortable speed $v_0$. We propose raising the absolute difference of speeds to an arbitrary power $c_e$ in order to further shape the behavior. $R_c$ makes agents avoid colliding with obstacles and with one another. $R_t$ is a commonly used reward component in goal-based RL environments, as it incentivizes the agents to reach their goal sooner rather than later.

\subsection{Energy and Metrics}\label{sec:energy-def}

When considering different reward functions, it is important to have metrics that are independent of the specific reward formulation in order to have a meaningful comparison. For this reason, we consider two types of metrics. 

Firstly, for each component of the reward function, we compute an unnormalized value measuring its performance, which can then be compared between different training runs. For example, we consider the total number of collision, regardless of the size of the collision penalty in the reward function.

% TODO: a few more citations? Other papers also use it
Secondly, we compute the mean energy expenditure of all agents in the scene. We use the following formula \citep{whittle_gait_2008, guy_pledestrians_2010}:
\[
    E_t = e_s + e_w v^2
\]
% Secondly, we compute the mean energy expenditure of all agents present in the scene, defined as $E_t = e_s + e_w v^2$ \citep{whittle_gait_2008, guy_pledestrians_2010},
where $E_t$ is energy spent per second, per kilogram body mass (J/kg$\cdot$s), $e_s$ and $e_w$ are constants, and $v$ is the current velocity (m/s). We use the values of $e_s = 2.23$ and $e_w = 1.26$ of a typical human. 
%We report energy values with a mass of $1$ kg. 
This induces an optimal walking speed of $v^* = \sqrt{\frac{e_s}{e_w}} = 1.33$ m/s which we use as the preferred walking speed. Work by \citet{bruneau_going_2015} suggests that when navigating to avoid collisions, people tend to choose a path that minimizes the energy usage, so we consider it to be a valuable metric describing the efficiency of the generated trajectory.

\subsection{Reward and Preferred Velocity}\label{sec:pref-vel}

\begin{figure}
\centering
\begin{subfigure}{0.22\textwidth}
    \includegraphics[width=\textwidth]{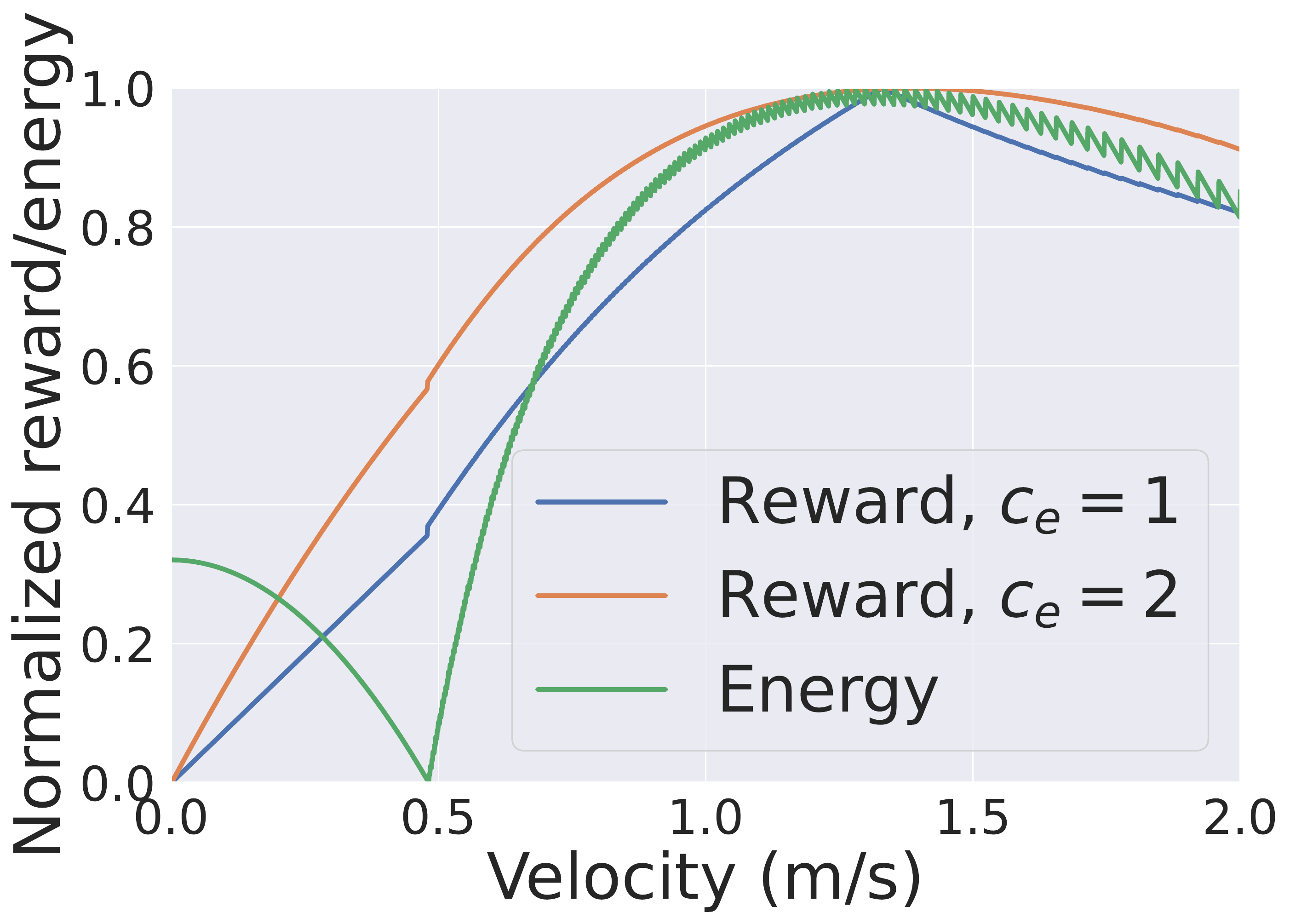}
    \caption{Reward and Energy.}
    \label{fig:reward-energy}
\end{subfigure}
\hfill
\begin{subfigure}{0.22\textwidth}
    \includegraphics[width=\textwidth]{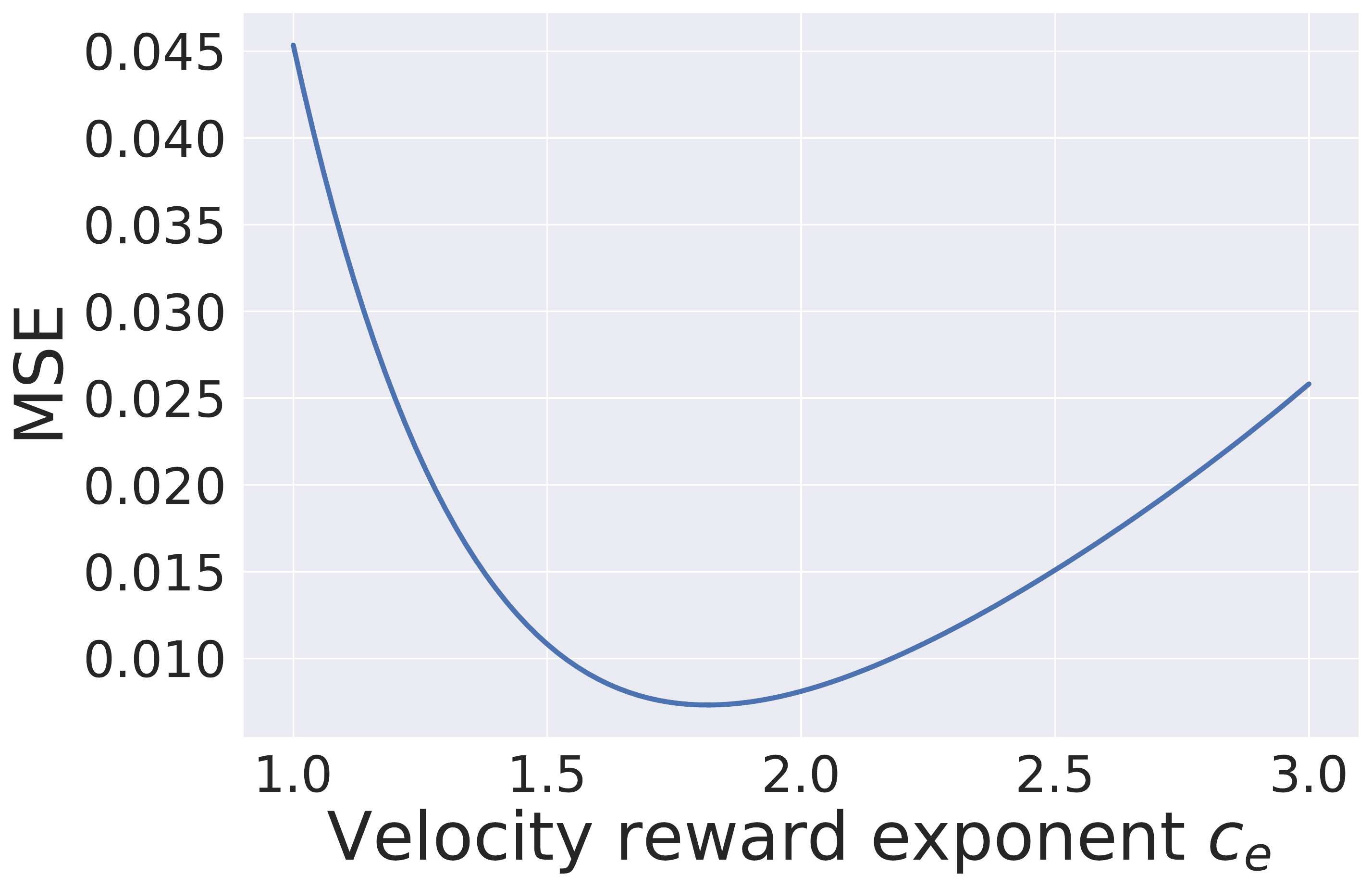}
    \caption{Velocity reward exponent.}
    \label{fig:reward-exponent}
\end{subfigure}
\hfill
\begin{subfigure}{0.22\textwidth}
    \includegraphics[width=\textwidth]{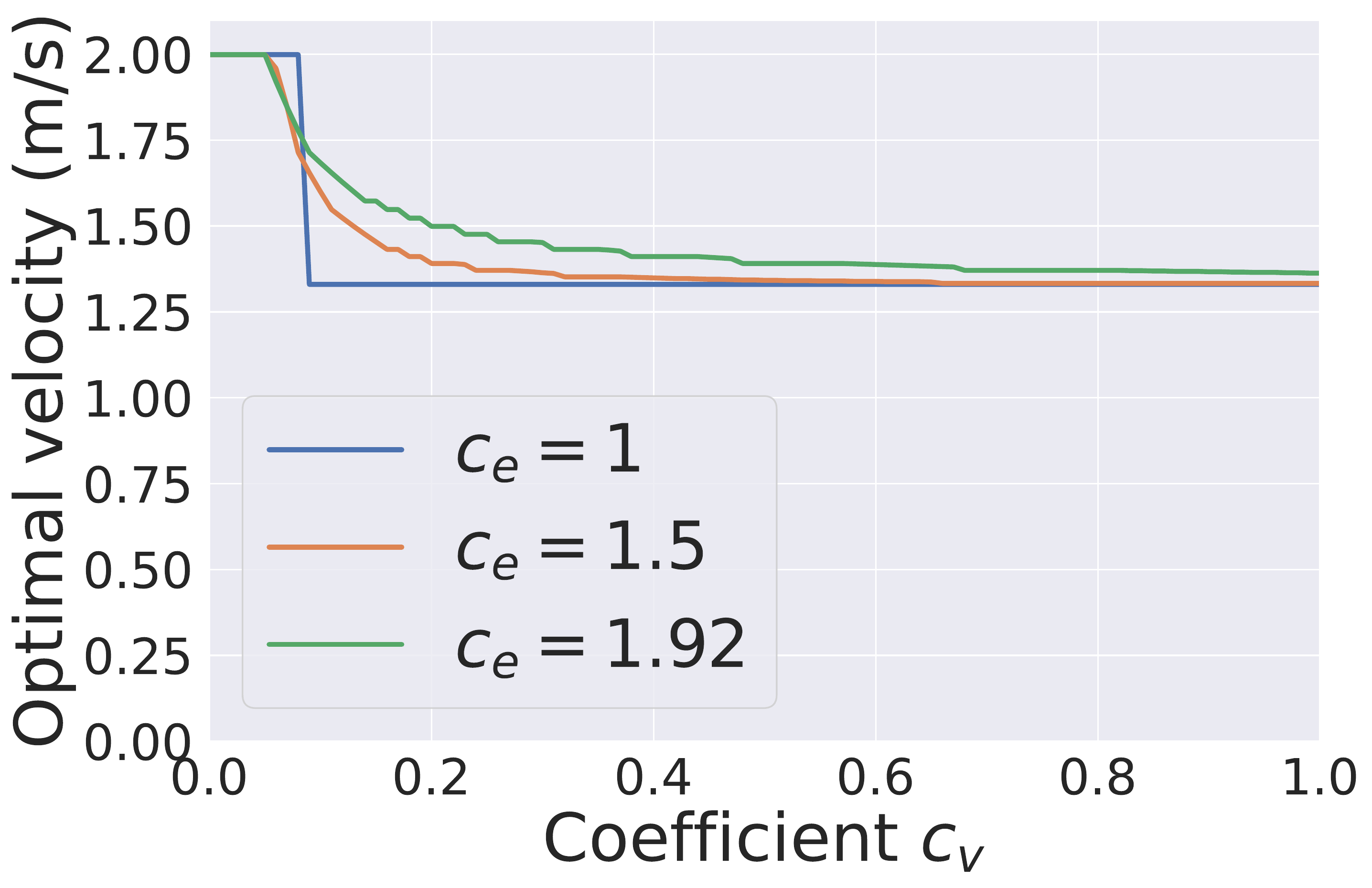}
    \caption{Velocity reward coefficient.}
    \label{fig:reward-velocity}
\end{subfigure}
\hfill
\begin{subfigure}{0.22\textwidth}
    \includegraphics[width=\textwidth]{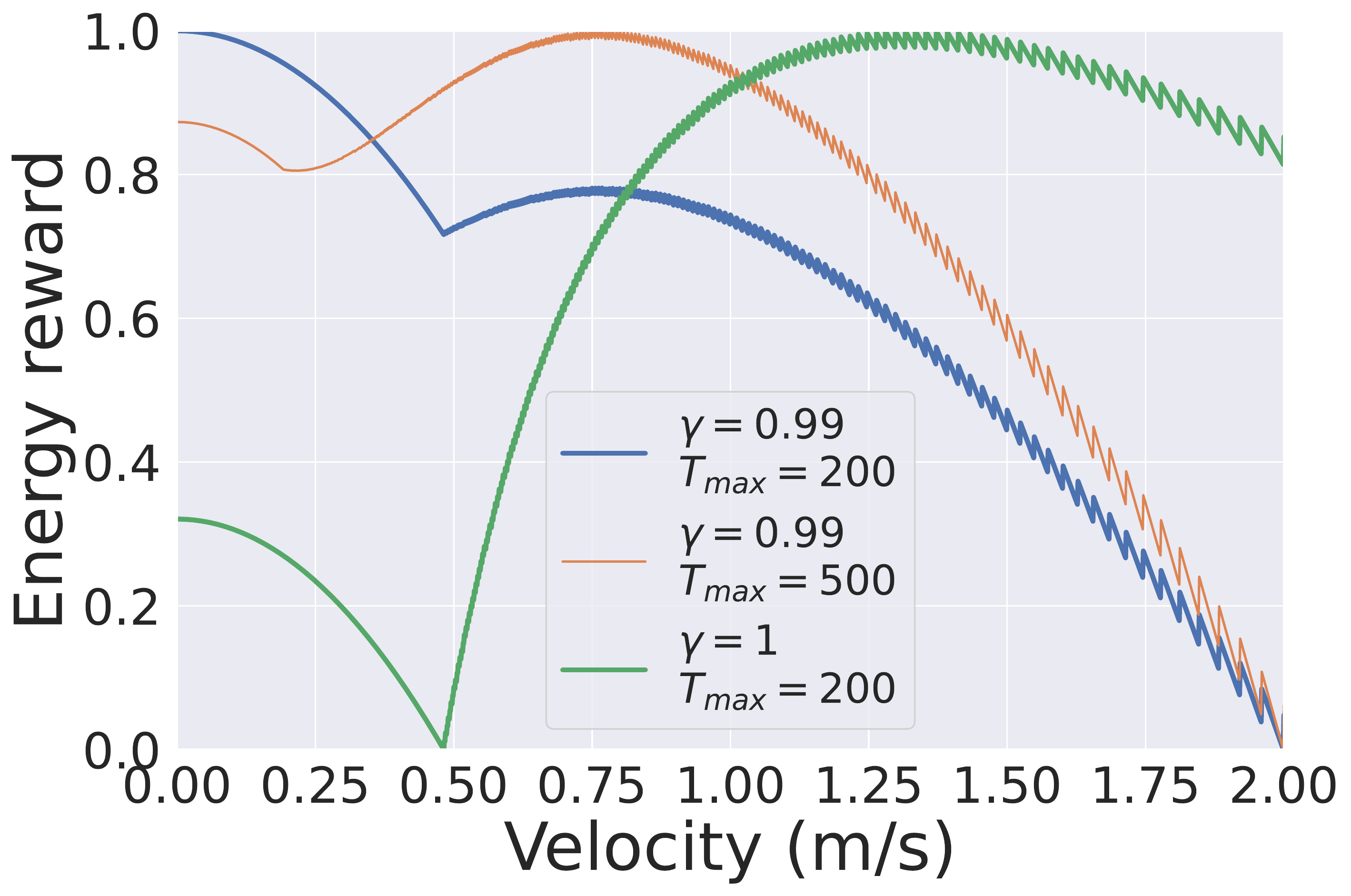}
    \caption{Energy-based reward.}
    \label{fig:energy-disc}
\end{subfigure}
        
\caption{\textbf{(a)} 
%The reward obtained by
Rewards and energy for an agent moving at a 
%given 
constant speed, in the simplified model described in Section~\ref{sec:pref-vel}.
%Section~\ref{sec:reward}.
All curves are normalized to be in the [0, 1] range in order to enable direct comparison. We consider energy values with the opposite sign, because energy is supposed to be minimized, while the reward is maximized. \textbf{(b)} MSE between the reward and the energy, as a function of the velocity reward exponent. \textbf{(c)} Optimal velocity as a function of the velocity reward coefficient $c_v$, varied by the exponent $c_e$. \textbf{(d)} Discounted negative energy expenditure as a function of velocity.}
\label{fig:reward-plots}
\end{figure}

As we see from the energy optimization mechanism, humans tend to move at a certain speed which is below their maximum possible speed. %Unless there is an emergency, we tend not to sprint everywhere, and opt for a more relaxed pace. 
This must be reflected in the reward function that the RL agent optimizes. However, it goes against the typical RL incentives to obtain rewards sooner due to the discounted utility model~\citep{sutton_reinforcement_2018}.

The interaction between the values of $c_g$, $c_p$, $c_v$, $c_e$, $c_t$, $\gamma$, and $T_{max}$, make the effective preferred velocity nontrivial to predict, and as a consequence -- to design. Consider a simplified environment model in which the agent must travel a distance $d$ towards its goal, with no other agents or obstacles. In this case, the only decision it makes is the velocity throughout its motion, under the assumption that it will travel to its goal in a straight line at a constant velocity. In this model, we can express the total obtained reward as:

\begin{equation}\label{eq:simple-reward}
    R(v) = \gamma^{T} c_g + \sum_{i=0}^{T} \gamma^i \left( c_p v \Delta t - c_v |v-v_0|^{c_e} - c_t \right)
\end{equation}
where $T = \min(T_{max}, \lceil \frac{d}{v\Delta t} \rceil) $
Due to the discounted sum whose bounds are dependent on the velocity itself, it is difficult to investigate this expression analytically. Nevertheless, we can gain some insight through numerically analyzing this model. Note, however, that this model does not capture the full complexity of RL optimization, and only serves to build general intuition.

Consider the following 
%basic 
set of parameters values 
%. These parameters have been 
-- based on prior work and then manually adjusted to produce reasonable behaviors -- %in order to serve 
as a starting point of our analysis:
$c_g = 10$, $c_p = 1$, $c_v = 0.75$, $c_e = 1$, $c_t = 0.005$ $v_0 = 1.33$, $\gamma = 0.99$, $T_{max} = 200$, $d = 8$, $\Delta t = \frac{1}{12}$.

Let us investigate the full reward as a function of velocity, alongside the negative energy expenditure as defined in Section~\ref{sec:energy-def}. This relation is shown in Figure~\ref{fig:reward-energy}. We can see that while there is a correlation between reward and energy, there are two main discrepancies. Firstly, the energy has a local optimum at $v=0$, which is caused by 
the velocities that cause agents to not reach their destination within the allotted time. Secondly, due to the absolute difference term in Equation~\ref{eq:simple-reward}, there is a sharp decrease in the reward (blue curve) which does not occur in the energy. 

To improve this, we evaluate the impact of the exponent $c_e$ from the reward function. In Figure~\ref{fig:reward-energy}, we also show $c_e = 2$ (orange). There, the curve is smoother and closer to the corresponding energy values when the energy is near the optimum. To quantify this, we vary the parameter $c_e$ and compute the mean square error between the two normalized curves in the range $1 \frac{m}{s} < v < 2 \frac{m}{s}$, as we consider lower velocities to be less relevant due to their low efficiency. As we show in Figure~\ref{fig:reward-exponent}, the optimal value under this simple model is $c_e = 1.92$. We further validate this in Section~\ref{sec:experiments}, where we use different values of $c_e$ for training actual RL agents. 

It is worth mentioning that with $\gamma = 0.99$, using $c_e = 2$ increases the effective optimal velocity to $v^* = 1.39 \frac{m}{s}$. This can be further adjusted using other parameters. While modifying $\gamma$ does not affect the optimal velocity when $c_e = 1$, any higher values of $c_e$ make it so that decreasing $\gamma$, increases $v^*$.

%\smallskip 
The other reward coefficients also have an impact on the optimal velocity. When adjusting $c_v$ with $c_e = 1$, there is a threshold around $c_v = 0.09$ below which the optimal velocity is the maximum value of $2 \frac{m}{s}$. Above that threshold, the optimal velocity is the preferred value of $v_0 = 1.33 \frac{m}{s}$. However, with $c_e > 1$, the transition between these two realms becomes more gradual, as we show in Figure~\ref{fig:reward-velocity}.

\subsection{Energy as reward} 

Finally, it is worth considering using energy directly as a reward function for training RL agents. At a glance, it seems like it would incentivize efficient motion at the optimal velocity. However, there are two apparent problems which arise in this paradigm. Firstly, as we already see in Figure~\ref{fig:reward-energy}, even in our simplified model there is an attractive local optimum at $v=0$. This is likely to be even more impactful in practical scenarios, because moving at the right speed would lead to a very low reward if the direction of the movement is wrong. The reason for this is the time limit present in the environment -- moving at a non-zero speed only pays off in terms of energy if the agent eventually reaches the goal. Otherwise, it will expend energy until the end of the episode, and by reducing its velocity, it can reduce the energy expenditure. Secondly, the commonly used method of reward discounting has a significant impact on the optimal policy. Using a discount factor of $\gamma = 0.99$ leads to a situation where the global optimum of the discounted energy-based reward function is standing still with $v=0$. Potential solutions to these problems include using a discount factor $\gamma = 1$ or a nonexponential discounting mechanism, increasing the time limit, and using a curriculum-based approach.

% Not sure if we should include this? Depends on the experiments
% To find the reward coefficients which provide the highest level of fidelity to the energy expenditure, we run a search on the entire parameter space of our model, with the objective being the MSE between normalized reward and energy curves for velocities $1 \frac{m}{s} < v < 2 \frac{m}{s}$. Interestingly, the best set of parameters has a preferred velocity value of $v_0 = 1 \frac{m}{s}$, and the remaining parameters increase the effective optimal velocity to $1.28 \frac{m}{s}$. In Section~\ref{sec:experiments} we show that this does indeed result in movement close to the desired velocity of $1.33 \frac{m}{s}$

\textbf{Conclusion.} Even in the absence of collision avoidance and other, more complex tasks, one must pay attention to the parameters defining the reward function. Notably, using an exponent in the velocity reward term causes other parameters to nontrivially affect the effective optimal velocity. For this reason, when designing the reward function, it is worthwhile to validate its parameters using a simpler model to ensure it has desired properties, whether that is closeness to the energy expenditure, or a specific value of the preferred velocity.
\section{Experimental setup}

\begin{figure}
\centering
\begin{subfigure}{0.11\textwidth}
    \includegraphics[width=\textwidth]{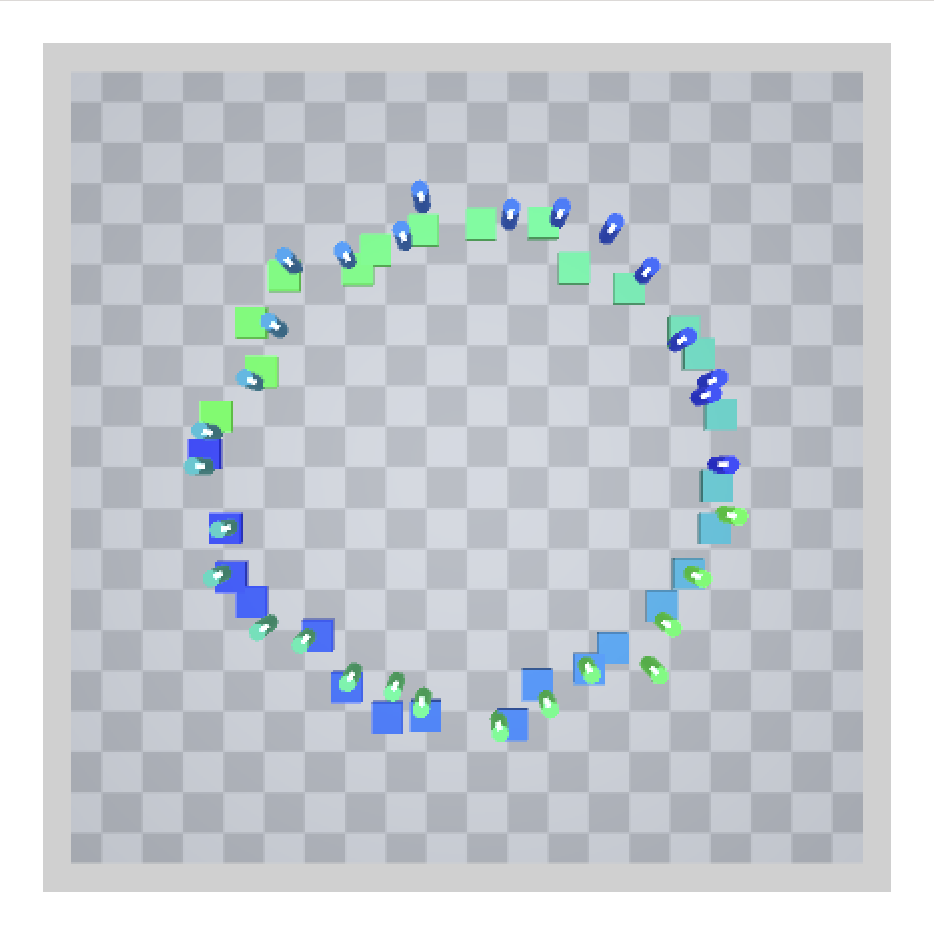}
    \caption{Circle scenario.}
    \label{fig:Circle}
\end{subfigure}
\hfill
\begin{subfigure}{0.11\textwidth}
    \includegraphics[width=\textwidth]{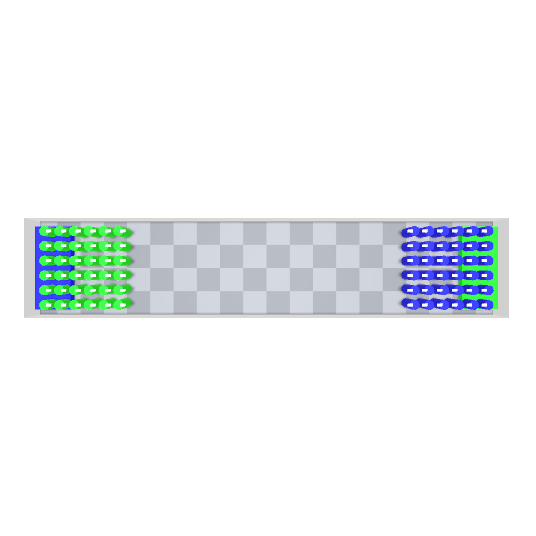}
    \caption{Corridor scenario.}
    \label{fig:Corridor}
\end{subfigure}
\hfill
\begin{subfigure}{0.11\textwidth}
    \includegraphics[width=\textwidth]{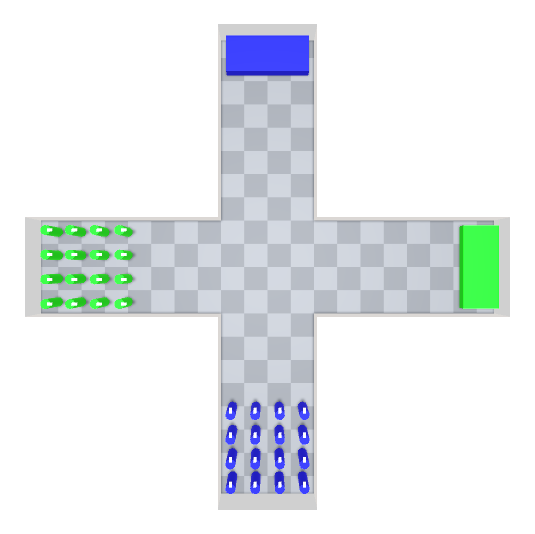}
    \caption{Crossing scenario.}
    \label{fig:Crossing}
\end{subfigure}
\hfill
\begin{subfigure}{0.11\textwidth}
    \includegraphics[width=\textwidth]{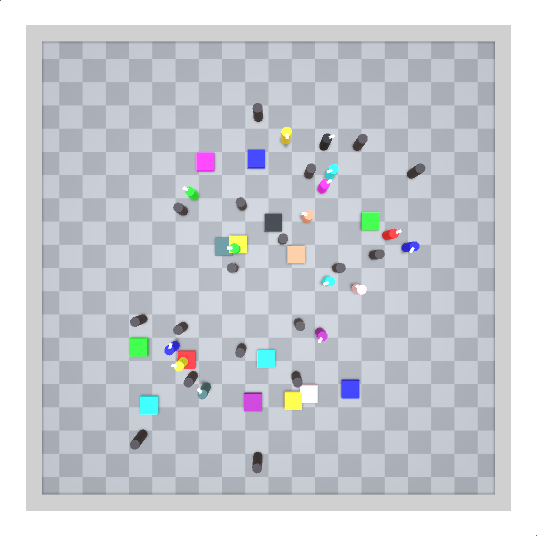}
    \caption{Random scenario.}
    \label{fig:Random}
\end{subfigure}
        
\caption{
%The initial positions of agents and their goals in different scenarios.
Agent's initial positions and goals in four scenarios:
\textbf{(a)} Circle with 30 agents. \textbf{(b)} Corridor with 72 agents. \textbf{(c)} Crossing with 32 agents. \textbf{(d)} Random with 15 agents.}
\label{fig:scenarios}
\end{figure}

In order to evaluate the impact and quality of the various design choices, we apply them on four commonly used crowd scenarios, in order to provide a wide range of interactions between agents: Circle, Corridor, Crossing, Random (see Figure~\ref{fig:scenarios}).
%. Each of these scenarios are shown in Figure~\ref{fig:scenarios}. 
In the Circle scenario, agents start on the perimeter of a circle, with a random noise applied independently in both Cartesian directions. Their goals are placed on the antipodal points of the circle, with an independent noise of the same magnitude applied. In Corridor, agents start at two ends of a straight corridor whose width is $4$ meters and length is $20$ meters. They start either in a regular grid or in a random formation, and their goal is to reach the opposite side of the corridor. In Crossing, the agents start at the ends of two corridors intersecting at a right angle, with the same size as in Corridor. Similarly, they spawn either in a regular grid or a random formation, and must reach the other end of their respective corridors. In Random, the agents' starting positions and goals are generated according to a uniform distribution with a given maximum size. In each of these scenarios, the area available to the agents is a square of 20x20 meters. In both Circle and Random, there are optional small obstacles placed randomly in the scene, represented as immovable agents.

\paragraph{Implementation}

The code\footnote{The code for the environment is available at \url{https://github.com/RedTachyon/CrowdAI}, and the training code is available at \url{https://github.com/RedTachyon/coltra-rl}.} used in this work is available online. 
We use identical agents represented as circles of radius $0.2$ m. Their collisions are treated as rigid body collisions, processed by the PhysX engine default in Unity 2021.3, used via the Unity Ml-Agents framework~\citep{juliani_unity_2020}.  We use a decision timestep $\Delta t = \frac{1}{12}$ s, similar to the values used in prior work. To obtain a more accurate simulation, the physics of the scene are updated 10 times after each decision, for an effective simulation timestep of $\frac{1}{120}$ s. The agent's action is repeated during each of these updates. With agent perception observations, each agent can see the 10 nearest agents.
%
%Each of these 
While these implementation details (\ie agent size, collision handling, physics engine, timestep) can also affect the resulting simulations and the training performance, we do not explore their impact in this work, because we expect it to be lower compared to the other choices listed in this paper. It is nevertheless important to be aware of these choices for reproducibility purposes. A single training takes between 1 and 5 hours on GPU, depending on the number of agents and the difficulty of the scene, while running 8 independent training runs in parallel. 

% TODO: add a figure with each scenario

\subsection{Policy Optimization}

In this work, we train RL agents using PPO with Generalized Advantage Estimation (GAE)~\cite{schulman_high-dimensional_2018} to estimate the advantages. The agents are trained in an independent paradigm with parameter sharing, that is they share the same policy network, but each agent takes its own action based on its private observations. The neural network outputs the mean of a Gaussian distribution, and the standard deviation is kept as a trainable parameter of the network.

\subsection{Network Architecture}\label{sec:architecture}

\begin{figure}
    \centering
    \includegraphics[width=1\linewidth]{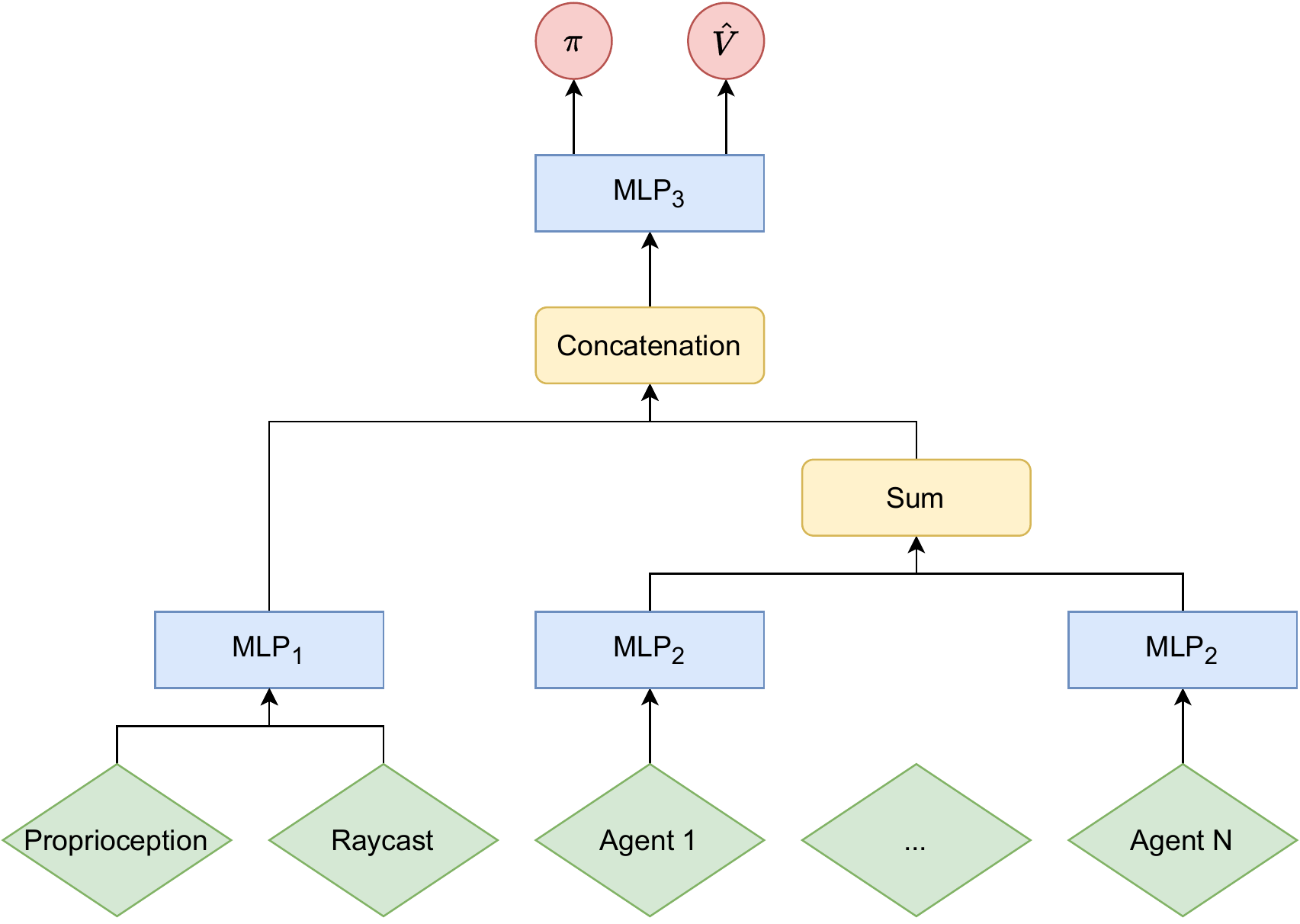}
    \caption{The neural architecture used as the policy. Green blocks represent inputs, blue blocks represent feed-forward neural networks, yellow blocks represent vector operations, red blocks represent outputs. Depending on the observation model used, certain elements of the architecture are disabled.}
    \label{fig:architecture}
\end{figure}

In order to appropriately process the Agent Perception observations, we use a neural architecture depicted in Figure~\ref{fig:architecture}. It is inspired by prior work such as Deep Sets~\cite{zaheer_deep_2017} and Mean Embedding~\cite{huttenrauch_deep_2019}, and extends the architecture used by \citet{xu_human-inspired_2021}. 

The main desirable property of our architecture is permutation invariance -- given multiple identical nearby agents, it should not matter in what order their representations are input into the network, as this order is completely arbitrary. Without this property, the agent would need to learn this invariance itself, which quickly becomes expensive as the number of observed agents grows. Furthermore, the architecture should be able to accept a variable number of observed agents, as this quantity will vary throughout the episode. 
For this reason, we use the following model as an embedding of nearby agents:
% \begin{align*}
\[
    \phi\left(\sum_i \psi(x_i)\right)
\]
% \end{align*}
% For this reason, we represent the nearby agents as $\phi\left(\sum_i \psi(x_i)\right)$,
where $\phi$ and $\psi$ are regular MLP neural networks, and $x_i$ is the observed information about an agent $i$. The summation is performed over all agents visible to the agent observing the scene. 
%In the absence of any visible agents, the embedding is a vector of zeros of the appropriate size.
Because of the summation operator, this architecture fulfills both previously stated desiderata, as the ordering information is lost, and any number of agents can be processed into a fixed-size embedding. This embedding is then concatenated with the main stream of the neural network, which processes the proprioceptive observations, as well as optionally the raycasting. 
%A full schematic of our architecture is shown in Figure~\ref{fig:architecture}.

\section{Experiments}\label{sec:experiments}

In this section we describe the specific experiments we performed, along with their results and interpretation. Experiments are primarily evaluated in terms of their obtained rewards and energy usage, but also other behavior characteristics when necessary.

\subsection{Dynamics and Observations performance}\label{sec:sweep}

% \begin{figure}
% \centering
% \includegraphics[width=\linewidth]{figs/experiments/reward-box.png}
        
% \caption{Comparison of training results in the 12 Circle scenario, with Agent Perception observations. }
% \label{fig:sweep-performance}
% \end{figure}

\begin{figure}
\centering
\begin{subfigure}{0.5\textwidth}
    \includegraphics[width=\textwidth]{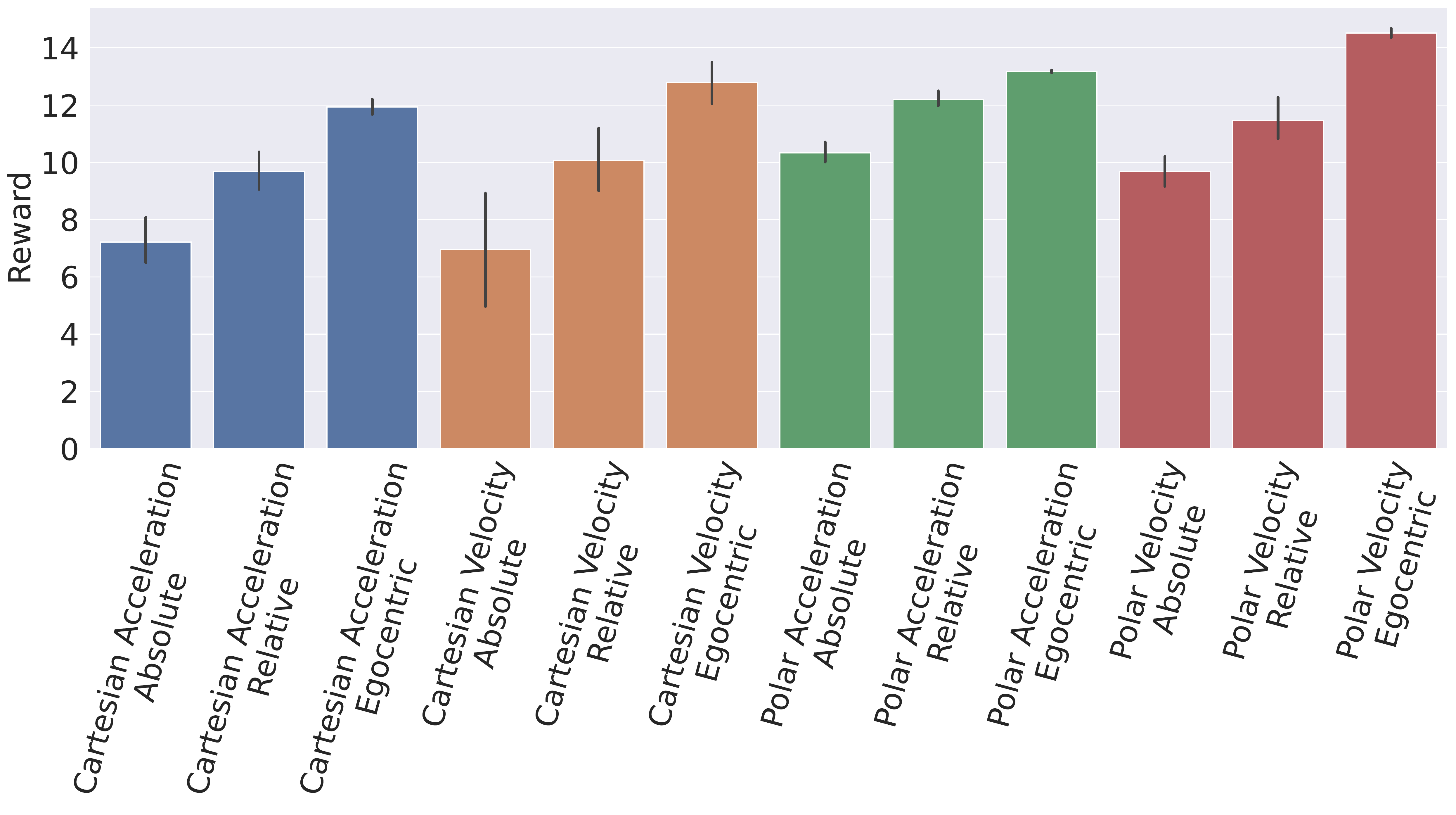}
    \caption{Reward function (higher is better).}
    \label{fig:sweep-reward}
\end{subfigure}
\hfill
\begin{subfigure}{0.5\textwidth}
    \includegraphics[width=\textwidth]{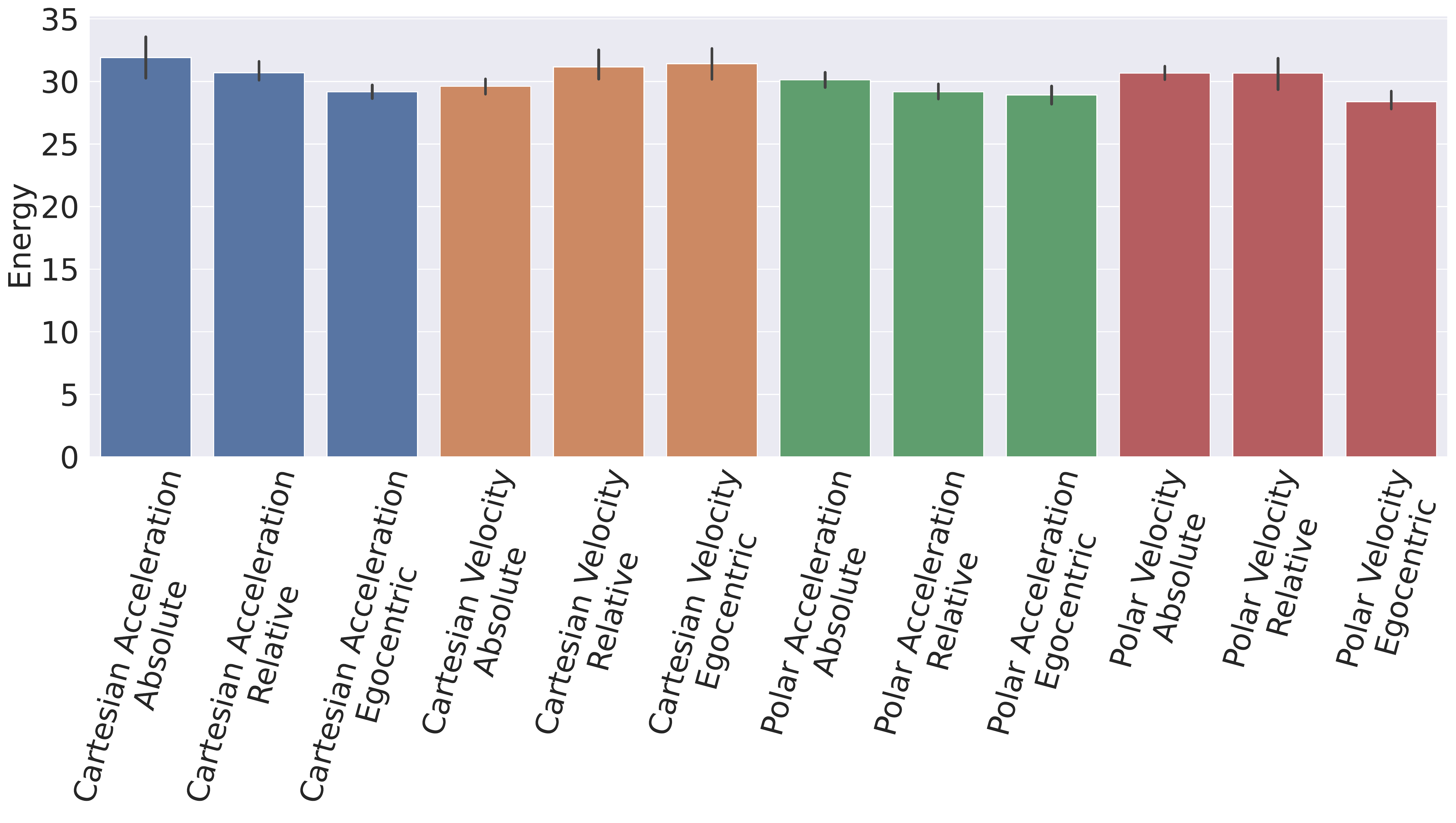}
    \caption{Energy expenditure (lower is better).}
    \label{fig:sweep-energy}
\end{subfigure}

\caption{Comparison of training results after a hyperparameter search in the Circle 12 scenario. \textbf{(a)} Mean episodic reward \textbf{(b)} Mean energy expenditure. Black bars represent the standard error of the mean.}
\label{fig:sweep}
\end{figure}

In order to robustly evaluate the performance of various dynamics and observation models, we run a hyperparameter search with each action space, with each observation reference frame, on a Circle scenario with 12 agents and a radius of 4 meters. Then, we use the best-performing hyperparameters of each model for further experiments used in this section. We sample 150 sets of hyperparameters for each (dynamics, observations) pair for a total of 1800 training runs. We use the default
%Tree-structured Parzen Estimator~\cite{bergstra_algorithms_2011} 
sampler available in Optuna~\citep{akiba_optuna_2019}. These runs use a fixed reward function, with parameters $c_g = 10$, $c_p = 1$, $c_v = 0.75$, $c_e = 1$, $c_c = 0.05$, $c_t = 0.005$. 

We show the results of the hyperparameter search in Figure~\ref{fig:sweep}. We consider the 5 best-performing hyperparameter sets for each model, and report their aggregate performance. There is a clear trend where agents with Egocentric observations perform better than Absolute and Relative versions. Similarly, Polar controls perform better than Cartesian controls, with Polar Velocity controls with Egocentric observations performing the best out of all investigated variants. Interestingly, in the case of Cartesian Velocity controls, while the reward follows the same trend, the energy usage is in fact the lowest with Absolute observations. This highlights the discrepancy between the reward function and the energy metric, showcasing the need for careful evaluation of emergent behaviors.

%We take the hyperparameter settings that achieved the highest rewards, and evaluate each of them on all scenarios with various observation and action spaces. 

\subsection{All Scenarios}\label{sec:experiments-other}

\begin{figure}
\centering
\begin{subfigure}{0.235\textwidth}
    \includegraphics[width=\textwidth]{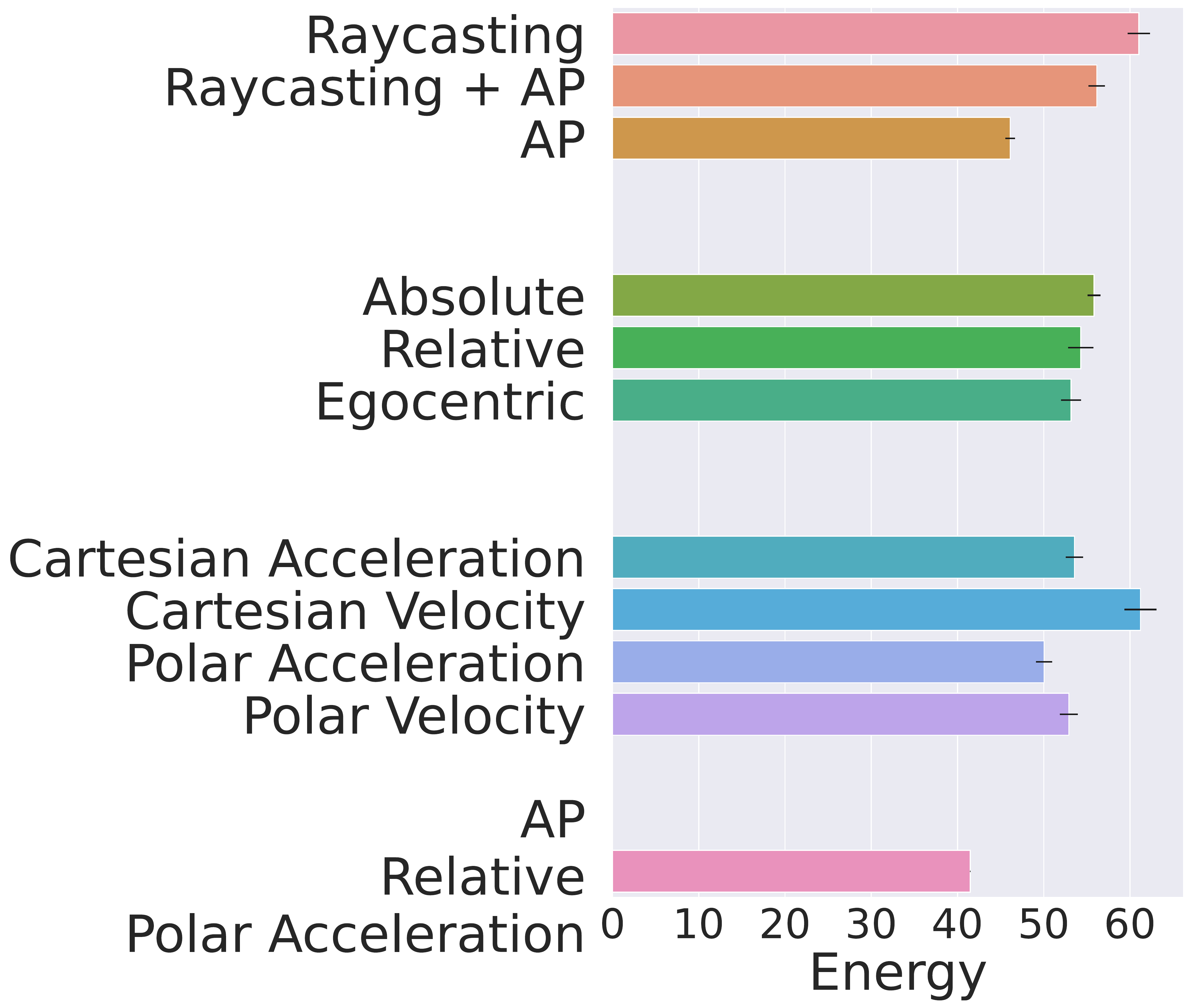}
    \caption{Scenario circle12}
    \label{fig:res-sce-circle}
\end{subfigure}
\hfill
\begin{subfigure}{0.235\textwidth}
    \includegraphics[width=\textwidth]{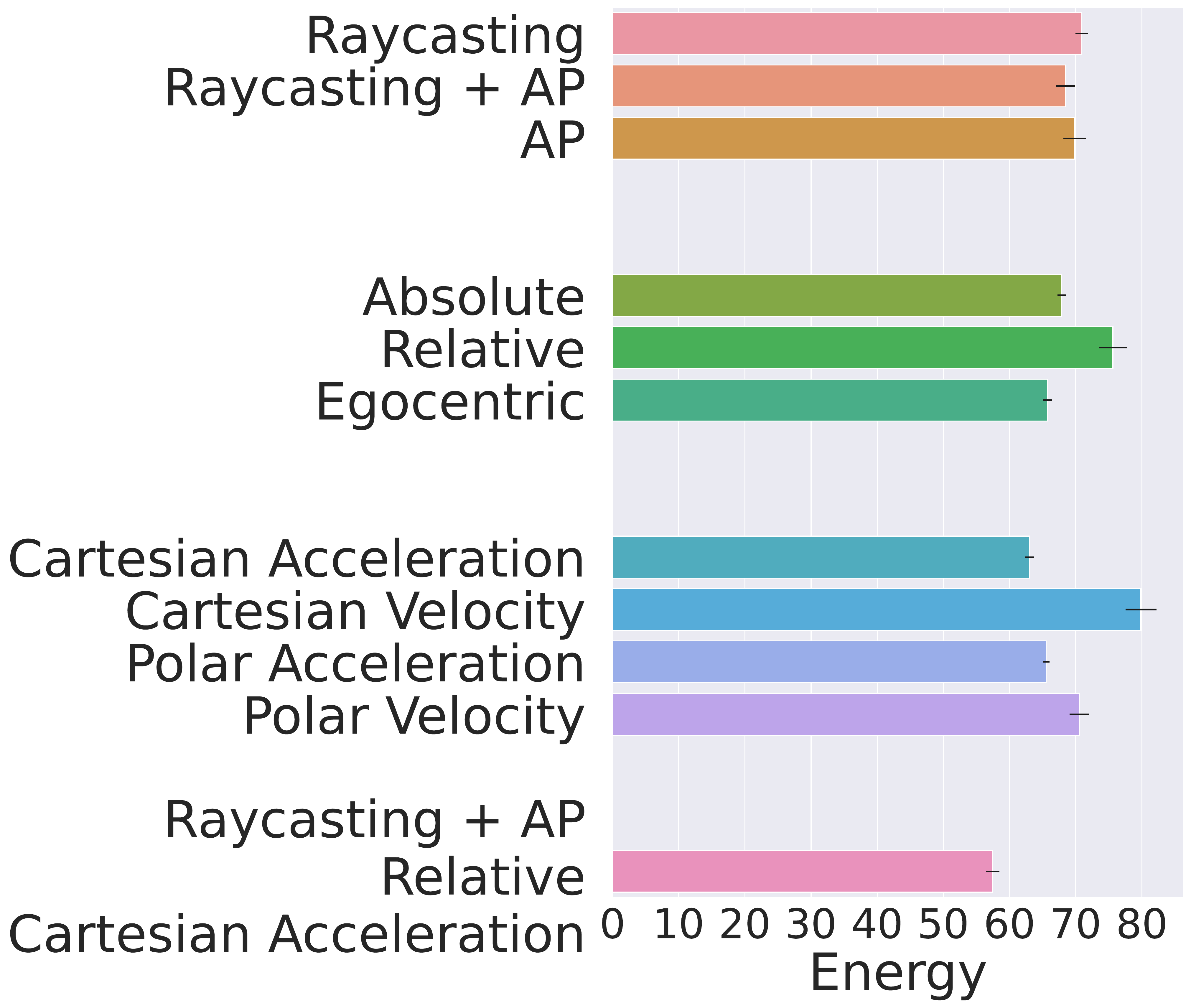}
    \caption{Scenario crossway50}
    \label{fig:res-sce-cross}
\end{subfigure}
\hfill
\begin{subfigure}{0.235\textwidth}
    \includegraphics[width=\textwidth]{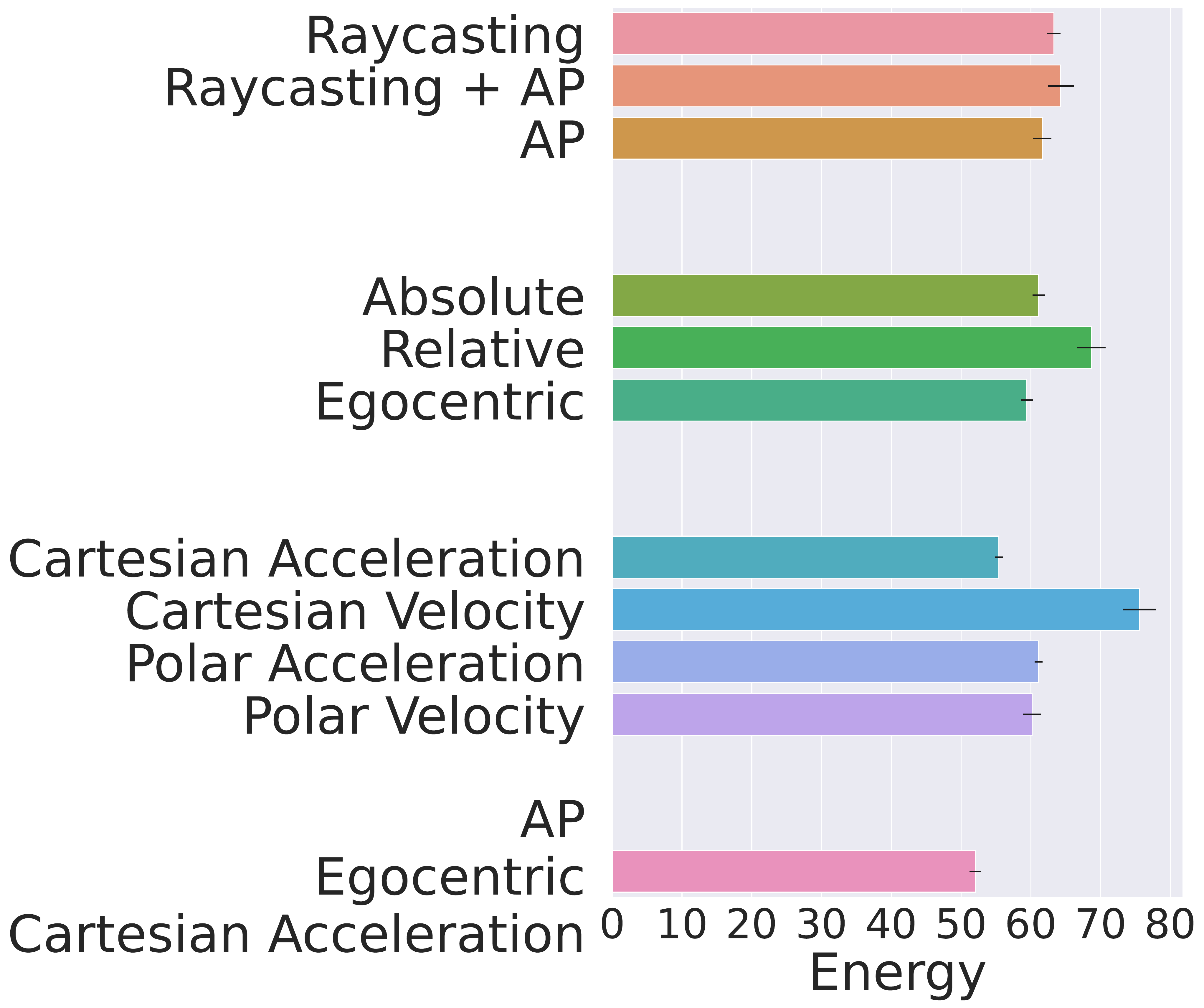}
    \caption{Scenario corridor50}
    \label{fig:res-sce-corr}
\end{subfigure}
\hfill
\begin{subfigure}{0.235\textwidth}
    \includegraphics[width=\textwidth]{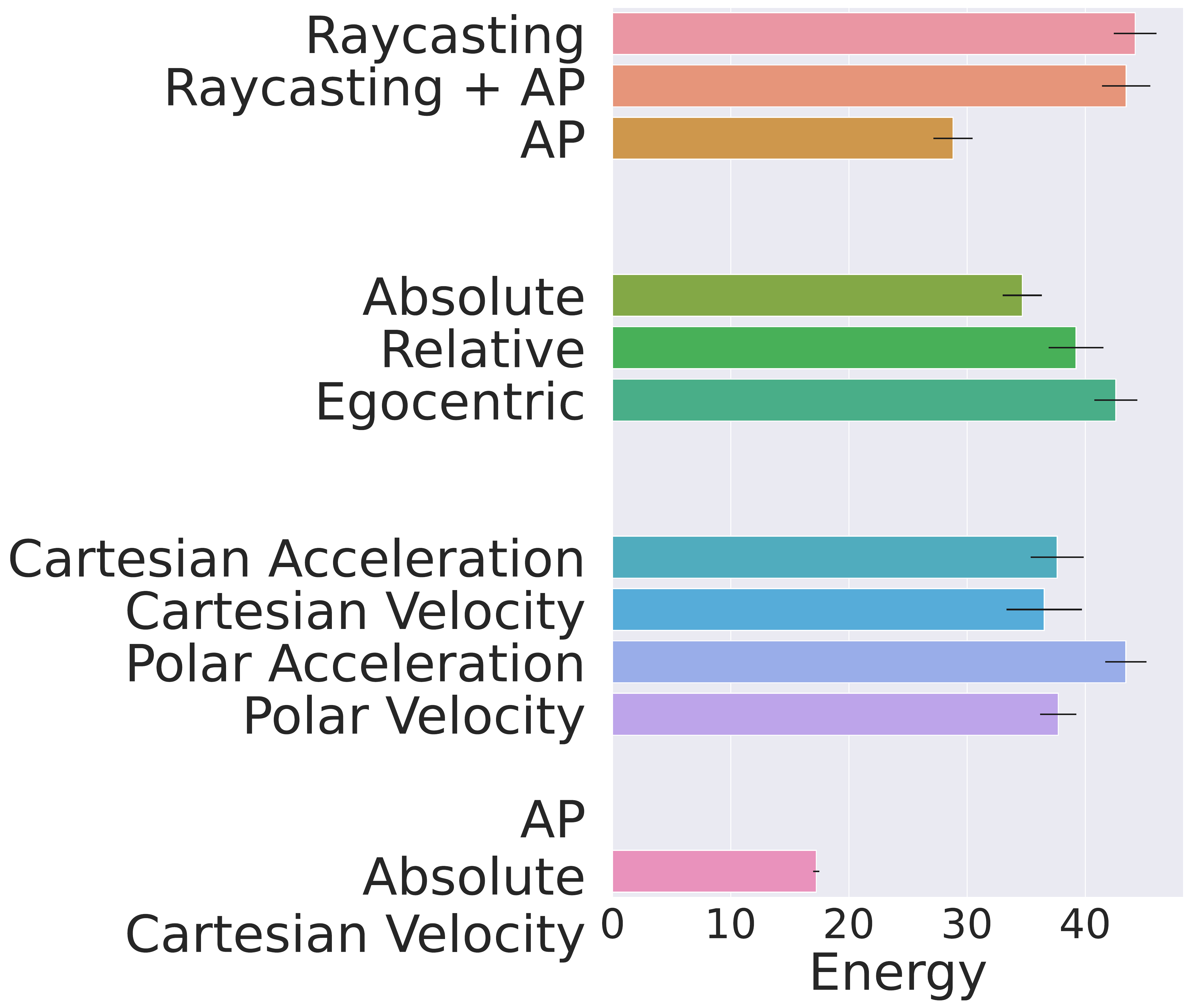}
    \caption{Scenario random20}
    \label{fig:res-sce-random}
\end{subfigure}
        
\caption{Comparison of various design choices in a given environment. The last bar corresponds to the best-performing agent across all design choices. All values are averaged across 8 independent training runs with different random seeds and otherwise identical parameters. Lower is better. AP stands for Agent Perception as defined in Section~\ref{sec:env-perception}.}
\label{fig:result-scenarios}
\end{figure}

Using the hyperparameters obtained in the experiment described in Section~\ref{sec:sweep}, we evaluate the various design choices on several environments. Specifically, we train agents in a Circle with 12 agents and a radius of 6 meters (circle12), a Corridor with 50 agents (corridor50), a Crossing with 50 agents (crossway50), and a Random scenario with 20 agents and random obstacles (random20).

Due to the large number of possible combinations (3 architectures $\times$ 4 dynamics $\times$ 3 observations $\times$ 4 scenarios), we consider the performance of each individual choice, averaging the remaining ones, in each environment separately. 
%We provide the full table of results in the supplemental material. 
The results are in Figure~\ref{fig:result-scenarios}. 

We run each training for 1000 PPO iterations to ensure sufficient time for convergence. Because the neural networks are relatively small (2-6 layers of 32-128 units, depending on the architecture and the model, found via hyperparameter optimization), the main bottleneck is the CPU power required to run many copies of the environment as opposed to the GPU memory.

While the exact results vary between scenarios, there are some regularities. Most notably, Agent Perception consistently outperforms raycasting-based approaches with otherwise well-performing settings. In the case of the crossway50 scenario, it is beneficial to include raycasting information around the surrounding walls, however overall it does not seem to provide a large benefit as compared to the pure AP approach. This is likely due to the static nature of the evaluated environments. Given the current position in the global frame, the agent can determine its proximity from obstacles. Unsurprisingly, including raycasting to perceive walls, deteriorates the performance in scenarios without a significant presence of walls, as this information effectively becomes an additional source of noise.

In most cases, it is best to use Egocentric or Relative observations, with some Relative runs sometimes performing the best. A notable exception is the random20 scenario, where a combination of Absolute observation with Cartesian Velocity dynamics outperforms any other option. The difference is the fact that in all other scenarios, the agents need to predominantly move in a general ``forward'' direction, whereas in the random scenario, the goal can be in an arbitrary position relative to the agent's orientation.

An interesting observation is the common ``failure mode'' of Raycasting models, particularly in Circle scenarios. They generally perform worse than Agent Perception models, but their qualitative behavior may be more desirable due to its asymmetry. Because they do not manage to reach the perfectly symmetrical trajectories achieved by Agent Perception, there is a higher variability in the individual trajectories, making them look more realistic. This indicates that simply training a strong RL algorithm on any objective which does not explicitly reward human-likeness is likely to lead to overly perfect, unrealistic trajectories.

\textbf{Conclusion.} We recommend using AP as opposed to the more commonly used raycasting for providing agents with the information about their surrounding. In scenarios where walls are a prominent feature, it may be beneficial to add raycasting which only perceives the distance to walls, and ignores other agents. In scenarios where agents may need to make sharp turns to reach their destination, Cartesian Velocity controls with Absolute observations are favorable. Otherwise, nonholonomic controls combined with egocentric or relative observations typically perform better. 

\subsection{Velocity Reward Exponent}

\begin{figure}
    \centering
    \begin{subfigure}{0.23\textwidth}
    \includegraphics[width=\textwidth]{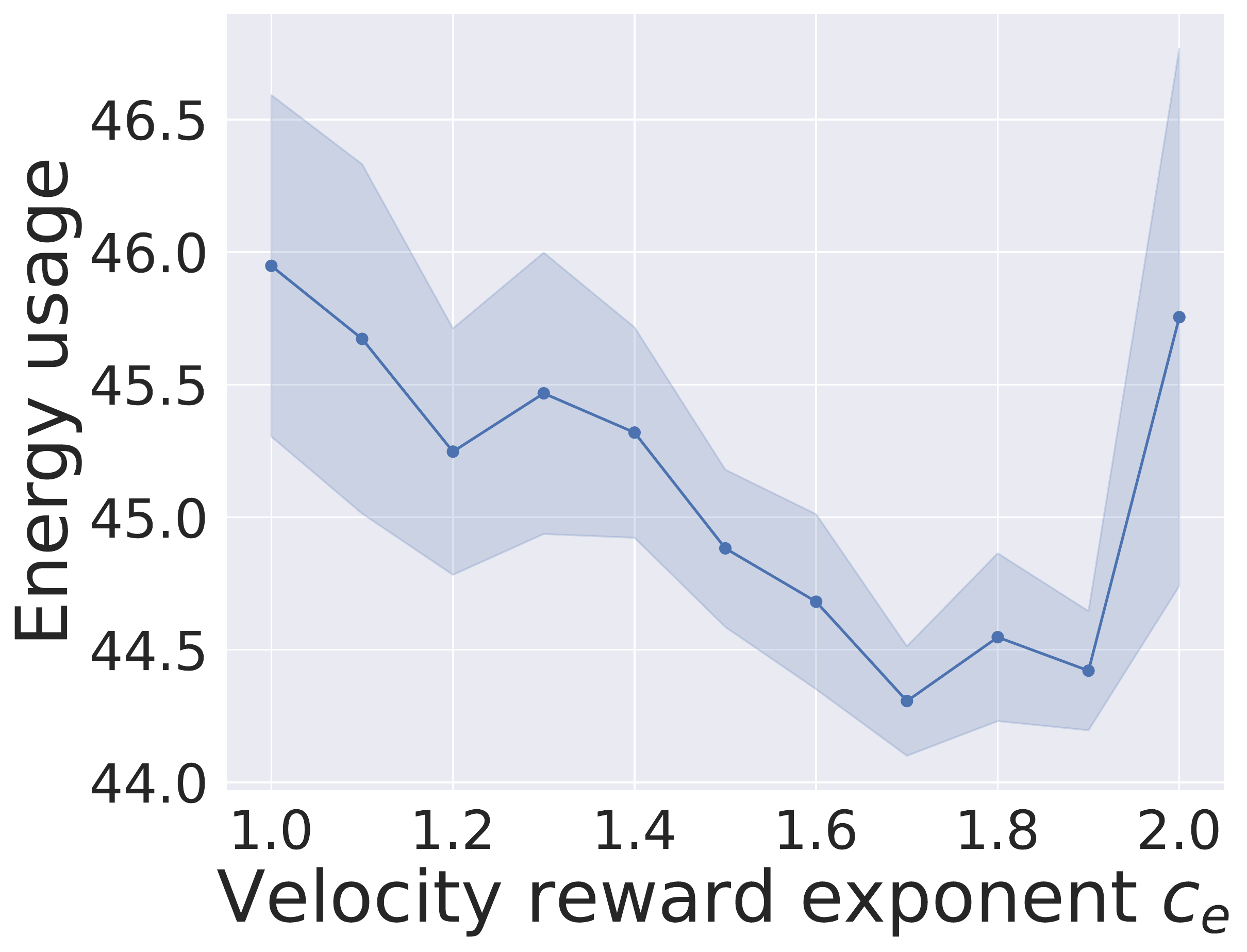}
    % \caption{Circle 30 scenario, AP Egocentric Polar Velocity}
    \caption{Circle 30 scenario}
    
    \label{fig:res-exp-circle}
\end{subfigure}
\hfill
\begin{subfigure}{0.23\textwidth}
    \includegraphics[width=\textwidth]{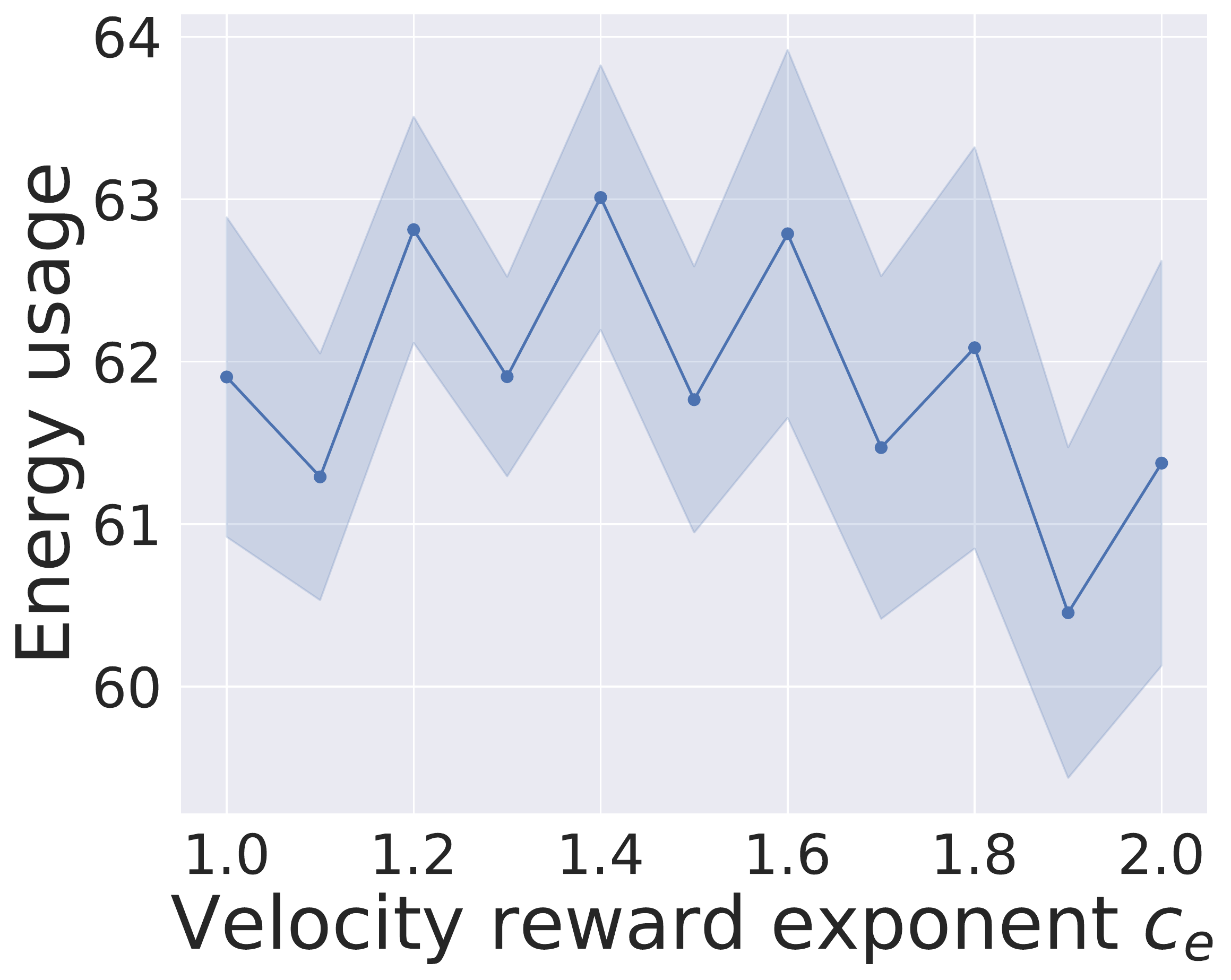}
    % \caption{Crossway 50 scenario, AP Egocentric Polar Acceleration}
    \caption{Crossway 50 scenario}

    \label{fig:res-exp-cross}
\end{subfigure}
    \caption{Comparison of energy usage in agents trained with a different exponent in the velocity term of the reward function. Lower is better. 
    % \textbf{(a)} Circle 30 scenario, AP Egocentric Polar Velocity model \textbf{(b)} Crossway 50 scenario, AP Egocentric Polar Acceleration model
    }
    \label{fig:results-exponent}
\end{figure}

In Section~\ref{sec:pref-vel}, we show that using an exponent in the velocity reward term makes it match more closely to the energy consumption. To validate this, we train Egocentric Polar Velocity agents on a Circle 30 scenario, and Egocentric Polar Acceleration agents on a Crossway 50 scenario. In both cases, an exponent $c_e > 1$ can lead to a higher efficiency in the trained agents, as compared to the simple $c_e = 1$ (see Figure~\ref{fig:results-exponent}). The exact optimal value of $c_e$ depends on the scenario and must be determined on a case-by-case basis.

% We average the results between 4 independent training runs and report the performance in Figure~\ref{fig:results-exponent}. With both model types, using $c_e = 1.92$ clearly leads to a lower energy usage, as well as an overall more stable performance. 

\subsection{Importance of collision penalty}\label{sec:reward-collision}

% \begin{figure}
%     \centering
%     \includegraphics[width=\linewidth]{figs/collision-circle.png}
%     \caption{The relation between the collision penalty magnitude, and the mean energy usage of trained agents.}
%     \label{fig:results-collision}
% \end{figure}

\begin{figure}
    \centering
    \begin{subfigure}{0.23\textwidth}
    \includegraphics[width=\textwidth]{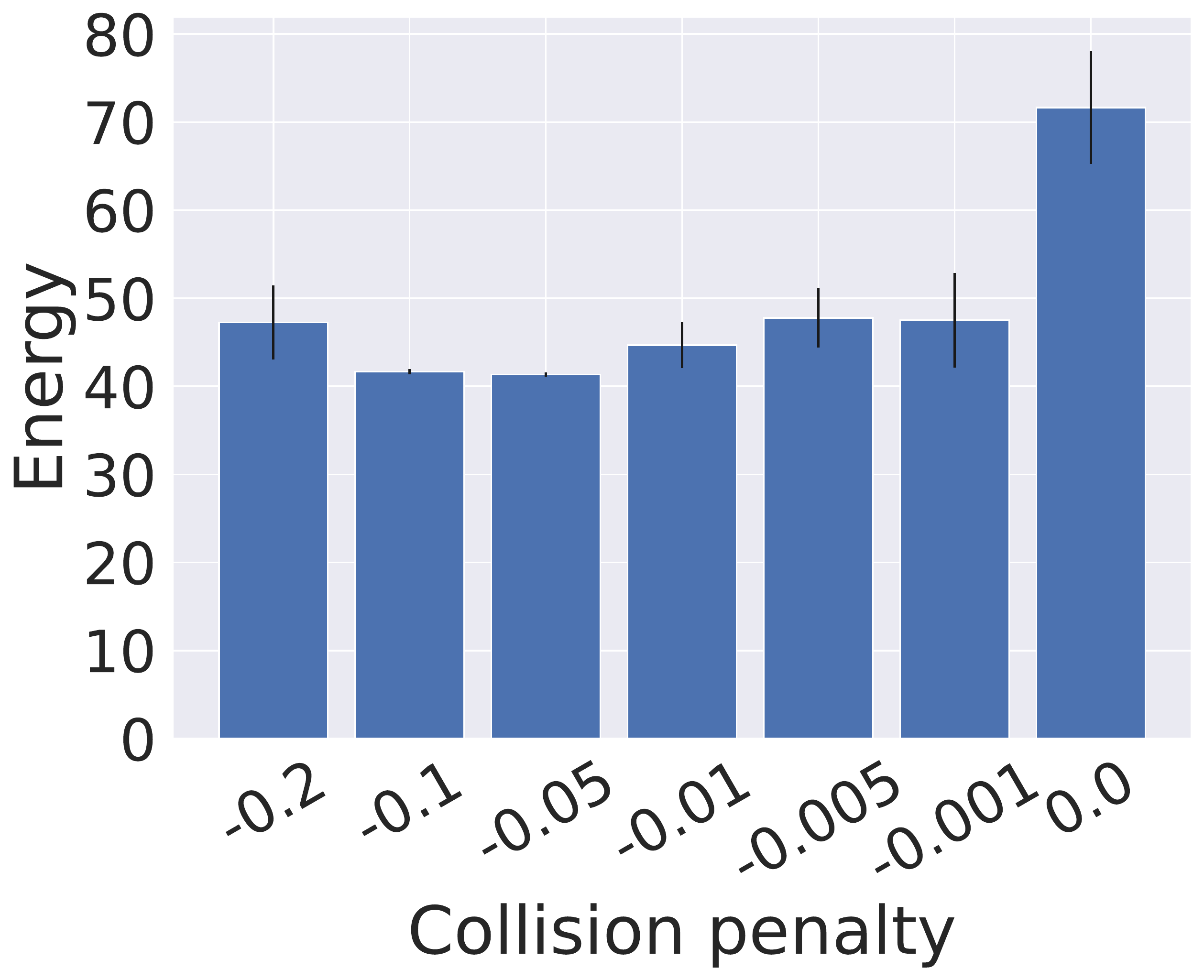}
    \caption{Energy usage.}
    \label{fig:res-col-circle-energy}
\end{subfigure}
\hfill
\begin{subfigure}{0.23\textwidth}
    \includegraphics[width=\textwidth]{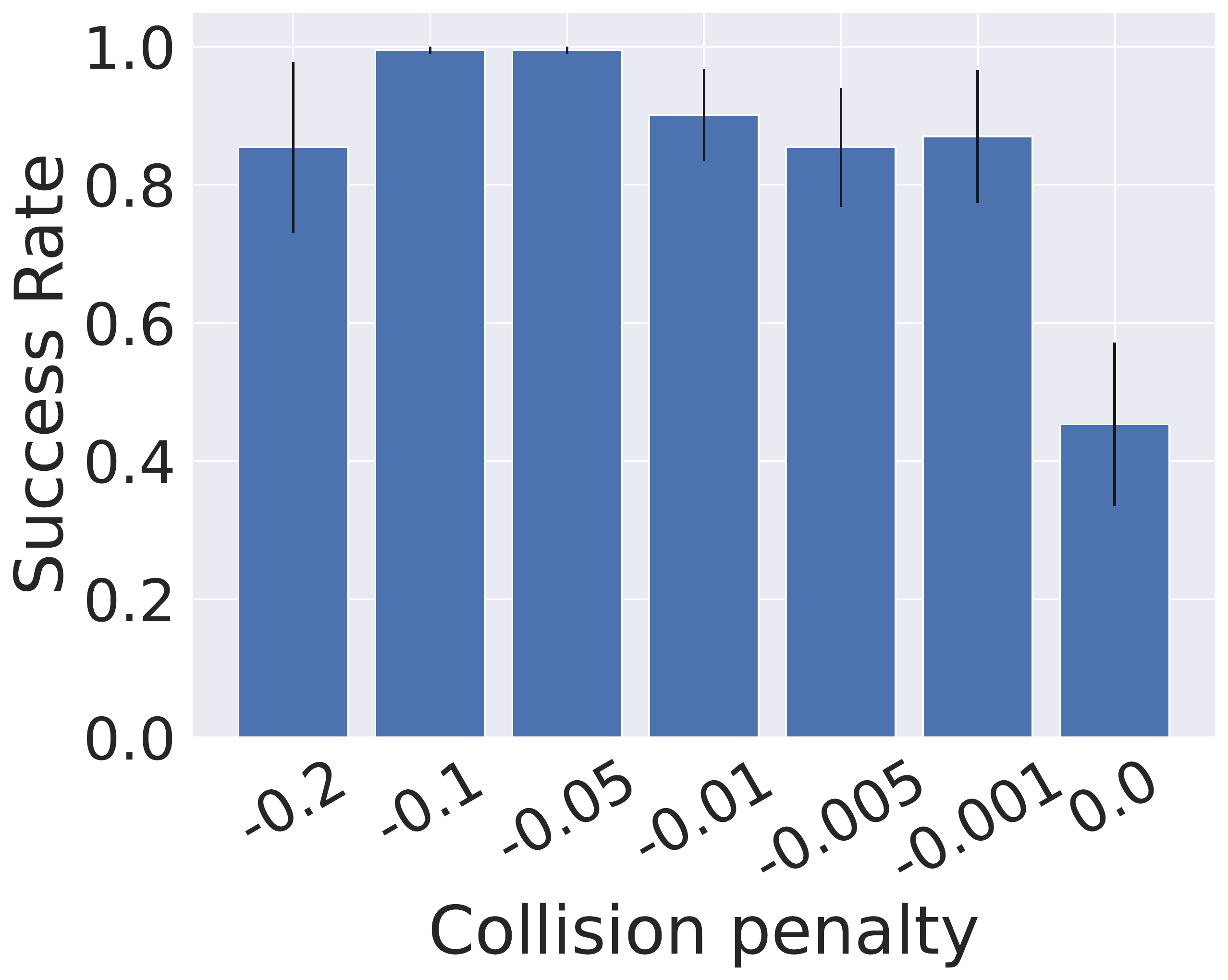}
    \caption{Success rate.}
    \label{fig:res-col-circle-success}
\end{subfigure}
    \caption{Comparison of energy usage and success rate in agents trained in Circle 12 scenario, with a varied collision penalty in the reward function.
    }
    \label{fig:results-collision}
\end{figure}

In a pursuit of simplicity in the design of the reward function, one might be tempted to eliminate the collision penalty altogether. After all, if collisions result in unfavorable physical results (\ie unexpected change of velocity), sufficiently intelligent agents should learn to avoid them by themselves, at least to the extent that is necessary for effective navigation.

In Figure~\ref{fig:results-collision} we show how the collision penalty affects the energy usage, and how often the agents reach their destination. There is an optimum around $-0.05$, where the agents reliably reach their goals. Using a penalty that is too high or too low leads to a deterioration of the agents' performance in terms of the navigation task.

\subsection{Common Failure Modes}

Due to the stochasticity inherent to RL training, the trained agents often exhibit various types of suboptimal behaviors. This can be identified via tracking the performance (in terms of the reward and energy), but also by observing emergent behaviors that the agents learn to execute. Here, we describe some of the common ways in which the RL-trained crowds are suboptimal.

\paragraph{Instability}

When training an RL agent using PPO, the trained policy is stochastic. This is required both for training, to ensure that the agent takes sufficiently diverse actions; but it is also at the core of the resulting policy. In order to deploy or evaluate the agent, we must choose a method of sampling actions from the output of the policy. The two natural options are taking the mean for the ``optimal'' action, or simply sampling from the distribution. The former does not fully correspond to the optimization objective, and in a crowd scenario, it can get stuck on an obstacle or another agent. The latter necessarily causes the policy to act randomly, which leads to potential unpredictable mistakes due to small erratic movements.

\paragraph{Value alignment failure}

\begin{figure}
    \centering
\begin{subfigure}{0.15\textwidth}
    \includegraphics[width=\textwidth]{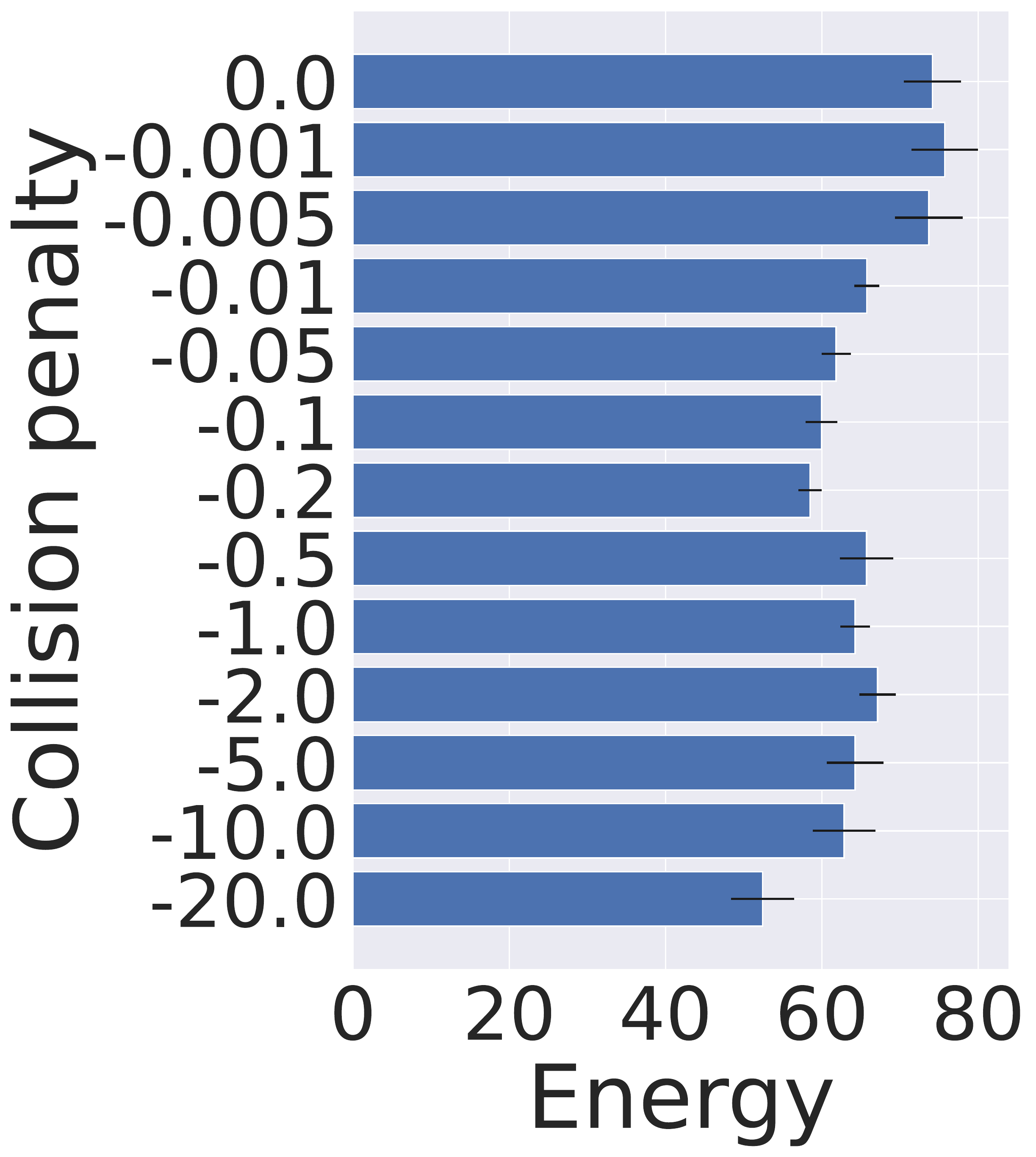}
    \caption{Energy usage.}
    \label{fig:res-col-cross-energy}
\end{subfigure}
\hfill
\begin{subfigure}{0.15\textwidth}
    \includegraphics[width=\textwidth]{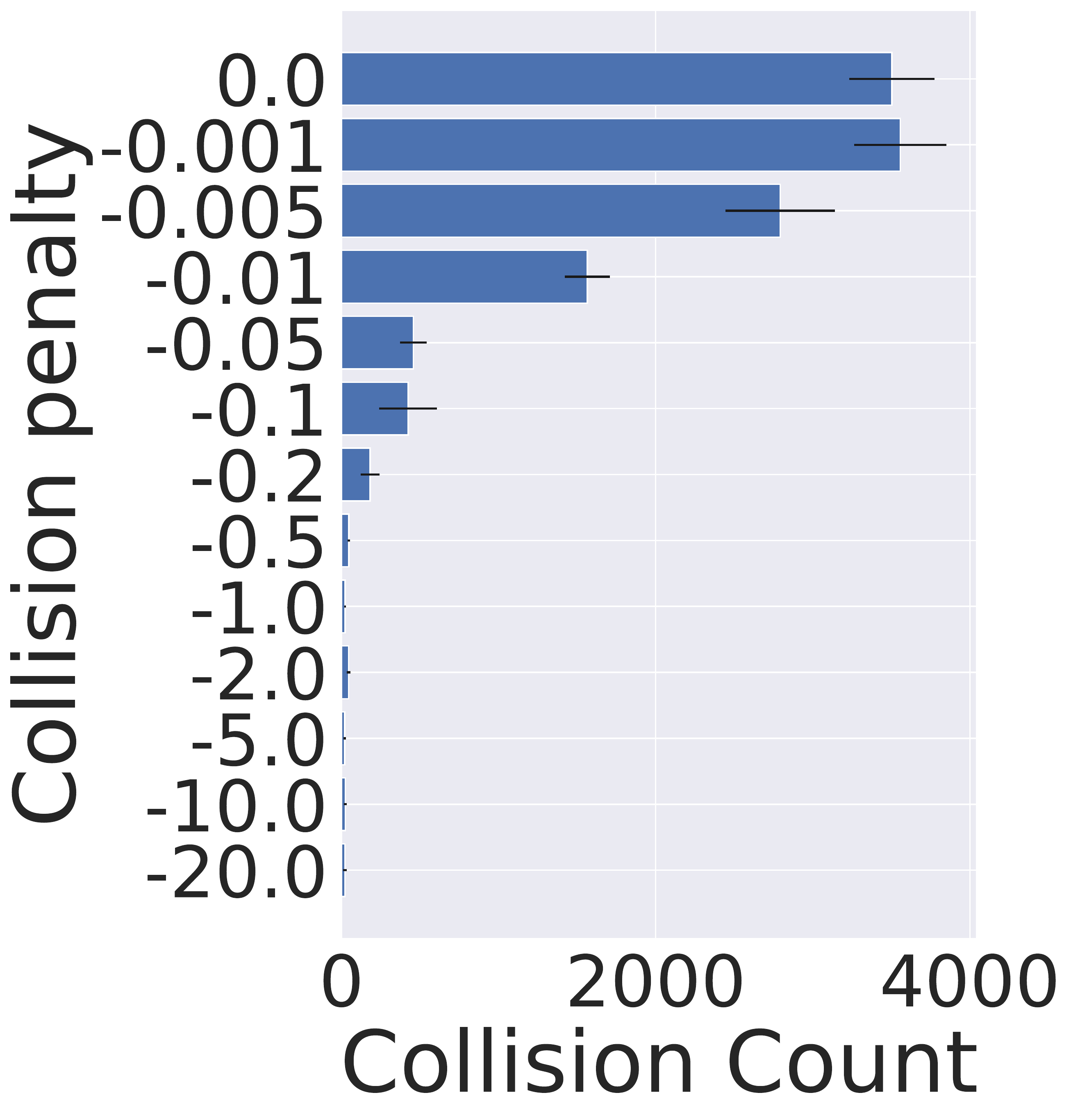}
    \caption{Collision count.}
    \label{fig:res-col-cross-success}
\end{subfigure}
\begin{subfigure}{0.15\textwidth}
    \includegraphics[width=\textwidth]{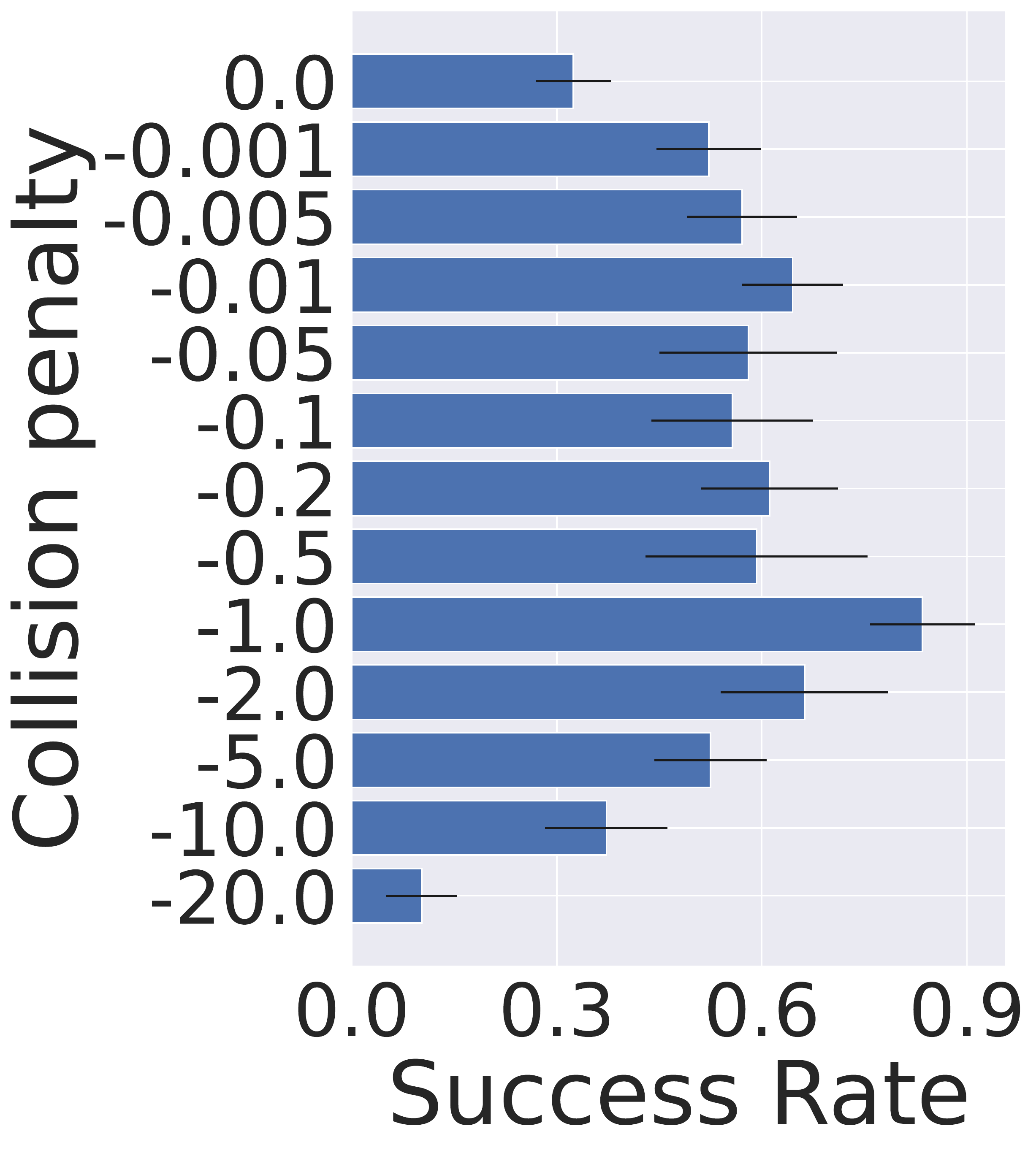}
    \caption{Success rate.}
    \label{fig:res-col-cross-finish}
\end{subfigure}
\hfill
    \caption{Comparison of energy usage, collision count and success rate in agents trained in Crossway 50 scenario, with a varied collision penalty in the reward function.
    }
    \label{fig:collision-value-alignment}
\end{figure}

% \begin{figure}
%     \centering
% \begin{subfigure}{0.5\textwidth}
%     \includegraphics[width=\textwidth]{figs/failure-energy.pdf}
%     \caption{Energy expenditure (lower is better).}
%     \label{fig:res-col-circle-energy}
% \end{subfigure}
% \hfill
% \begin{subfigure}{0.5\textwidth}
%     \includegraphics[width=\textwidth]{figs/failure-collision_count.pdf}
%     \caption{Collision count (lower is better).}
%     \label{fig:res-col-circle-success}
% \end{subfigure}
% \begin{subfigure}{0.5\textwidth}
%     \includegraphics[width=\textwidth]{figs/failure-finish_rate.pdf}
%     \caption{Success rate (higher is better).}
%     \label{fig:res-col-circle-energy}
% \end{subfigure}
% \hfill
%     \caption{Comparison of energy usage, collision count and success rate in agents trained in Crossway 50 scenario, with a varied collision penalty in the reward function.
%     }
%     \label{fig:collision-value-alignment}
% \end{figure}

As we describe in Section~\ref{sec:reward}, the reward function used in this and prior work is a weighted sum of several components with largely arbitrary coefficients. This implies that there is no guarantee it will correspond well to the actual objective we intend the agents to achieve. Furthermore, while the energy usage can be a useful, non-arbitrary metric, it is also sensitive to some details of the practical RL setup, as we show next.

Consider the experiment described in Section~\ref{sec:reward-collision}, investigating the importance of the collision penalty. We perform the same analysis on the Crossway 50 scenario with Polar Velocity dynamics and Egocentric observations, extending the range of evaluated collision penalties to $-20$. We present these results in Figure~\ref{fig:collision-value-alignment}. 

When considering the energy usage and collisions, the result seems to follow the intuition with a collision penalty up to $-1$. When the collision penalty reaches $-20$, we expect the performance to deteriorate, since the penalty is so steep that the agents will never put themselves at any risk of collision. We confirm this by investigating the success rate, which decreases with large collision penalties. However, at the same time, the energy usage also decreases -- in fact, using a very large collision penalty leads to the lowest energy usage among the evaluated options. So while energy has its value as a metric, it clearly breaks down in extreme cases, where most agents remain stationary.

This phenomenon is caused by a relatively short time limit. In general, energy minimization induces a simple optimal velocity (see Section~\ref{sec:energy-def}), in RL we must set a time limit after which the episode is terminated. The energy penalty for not finishing the episode is relatively low when compared to the energy cost of actually navigating to the destination, and an undesired emergent behavior may turn out to be optimal according to the metrics. 

Depending on the design choices made, this same problem can manifest itself differently. When training on the same setup, but using different design choices, a commonly occurring emergent behavior is that one group of agents efficiently navigates to their destination, while the other one remains stationary. Similarly to the previous situation, the energy usage metric is very low in that situation, but the success rate plateaus at 50\%.

\section{Discussion}

In this work, we present an analysis of various design choices made by designers of RL-trained crowd simulation systems. We show that many of these choices, typically ignored by researchers, can in fact significantly impact the resulting simulation, in particular when evaluated in terms of the energy efficiency. Here we summarize the main findings of this paper.

We show that the commonly used raycasting underperforms when compared to a method we call Agent Perception where the information about neighboring agents is directly available. This is likely a consequence of a much simpler and accessible representation of that information. Even when raycasting uses frame stacking which enables movement perception, the reasoning needed to infer the positions and velocities of nearby agents is rather complex.

Designing the right reward function is also important for obtaining desired properties of the motion. Navigation, speed control, and collision avoidance rewards, all have to balanced in just the right way that each of them contributes to the agent's decisions. Crucially, quantitative comparisons of crowds are non-trivial, because even ignoring the question of believability, the arbitrary reward function and energy usage are flawed in certain scenarios.

Qualitative properties of the obtained motion in the Circle scenario indicate that a naive approach of reward maximization for any reward that does not explicitly incentivize human-like behavior is likely to create trajectories that look artificial, or even too carefully choreographed to pass for natural human behavior. While decreasing the model's capacity and performance might lead to more believable behavior in the short term, we believe a more deliberate approach is necessary to truly approach human-like behavior.

In summary, our main findings are as follows:
\begin{enumerate}
    \item Direct agent perception outperforms simple raycasting
    \item Egocentric controls tend to outperform absolute ones
    \item The reward design is important and nontrivial
    \item Many failure modes may still occur in RL trained crowds
    \item Simple reward is not sufficient for human-like behavior
    
\end{enumerate}

\subsection{Limitations and Future Work}

All experiments in this work are performed on relatively small, static scenarios with a single destination. The described design choices mostly affect local navigation, and more complex scenarios can be expressed as a sequence of partial objectives or checkpoints. That being said, naively implementing this would likely cause issues near the transition points where agents switch their destination. Therefore, a more complex training scenario would be beneficial so that the agent is exposed to these situations.

Furthermore, each agent is only trained on a single scenario. Prior work suggests that using various scenarios in the training process enables generalization, which was considered out of scope of this work. We also limit our analysis to the efficiency of the resulting trajectories, ignoring realism or believability. 

% \subsection{Future Work}

% There are many potential ways to extend this work and improve the performance of RL-trained crowds. It is likely that each of the design choices investigated in this work can be improved upon. 

Similarly to prior work, we train agents using an arbitrarily-designed reward function. While using the energy usage as a reward has certain problems (see Section~\ref{sec:pref-vel}), it might be a viable option by using a curriculum-based approach where the reward function changes as the training progresses; and a different discounting mechanism that improves the global optimization properties. 
Furthermore, by using recent work on evaluating the realism of generated trajectories~\citep{daniel_perceptually-validated_2021}, 
% Furthermore, by using an external measure of the realism of a trajectory,  
a promising direction is using a realism metric as a reward. This would allow going beyond efficiency, and creating crowds which behave in a believable way.

It is also possible to improve the dynamics available to the agent. In this work, we use relatively simple, 2-DoF models, but the RL paradigm makes it viable to implement arbitrary nonholonomic constraints like sidesteps or walking backwards, without having to change the learning logic. Thus, a promising option is introducing a more complex, human-like range of motion actions available to the agent, with the goal of improving the believability of motion. 

% Finally, adding more of a structure in the relations between individual agents would enable modelling more interesting, life-like scenarios. If a few agents within the crowd happen to be a family, it would make sense for them to prefer staying near one another. This could also enable a more hierarchical approach to simulating big scenes, where the crowd is composed of smaller groups, and each of those units consists of several agents which follow the same rules as the groups. The potential benefits of this approach include better scalability, ability to simulate at various scales, potentially introducing a ``level of detail'' setting, and better coordination between large amounts of agents.

\subsection{Conclusions}

% We evaluated various design choices on multiple crowd scenarios, and compared them in terms of the reward function and their energy efficiency. We show that, as a general rule, egocentric observations combined with non-holonomic dynamics outperform absolute observations with holonomic dynamics. We also formally evaluate a way for agents to directly observe other agents, and show that it performs better than the more commonly used raycasting. Finally, we highlight some of the concerns when designing a reward function for collision avoidance and setting a preferred velocity.

Crowd simulation with RL is a complex problem, and despite recent advances, many challenges remain. Observations, actions, underlying physics, and especially the reward function, all have a significant impact on the results, and a lack of attention to these design choices makes it impossible to compare various approaches. We bring these issues to attention, and introduce a basic methodology for comparison between various approaches by comparing the energy expenditure under a specified time limit. In the absence of a standard benchmark, we call researchers to be more explicit and precise about their choices, and encourage them to explore different options in their work, to ensure robustness of their approach.

\section*{Acknowledgments}
This work has received funding from the European Union’s Horizon 2020 research and innovation programme under the Marie Skłodowska-Curie grant agreement No 860768 (CLIPE project). This work was performed using HPC resources from GENCI–IDRIS (Grant 2022-AD011013684). This work was also supported by the Hi!Paris institute of IP Paris, through the CREATIVE AI fellowship.

% \section*{References}

% Please ensure that every reference cited in the text is also present in
% the reference list (and vice versa).

% \section*{\itshape Reference style}

% Text: All citations in the text should refer to:
% \begin{enumerate}
% \item Single author: the author's name (without initials, unless there
% is ambiguity) and the year of publication;
% \item Two authors: both authors' names and the year of publication;
% \item Three or more authors: first author's name followed by `et al.'
% and the year of publication.
% \end{enumerate}
% Citations may be made directly (or parenthetically). Groups of
% references should be listed first alphabetically, then chronologically.

%%Vancouver style references.
\bibliographystyle{cag-num-names}
\bibliography{text/references}

\newpage

\section*{Appendix}

Here we provide the training plots of the experiments described in Section~\ref{sec:experiments-other}. For each scenario we plot together the three model types (Direct Agent Perception, Raycasting and both) together, for each combination of observations and dynamics. For each experiment, we show both a plot with the full range of rewards, as well as a version zoomed in to the high range of rewards obtained in at least some of the experiments. We also show the energy expenditure, where lower values correspond to better policies. Each line is averaged across 8 independent runs which differ only by the random seed. The shaded area corresponds to the standard error of the mean. Note that the rewards cannot be directly compared between different scenarios.

\begin{figure*}
    \centering
    \includegraphics[width=\linewidth]{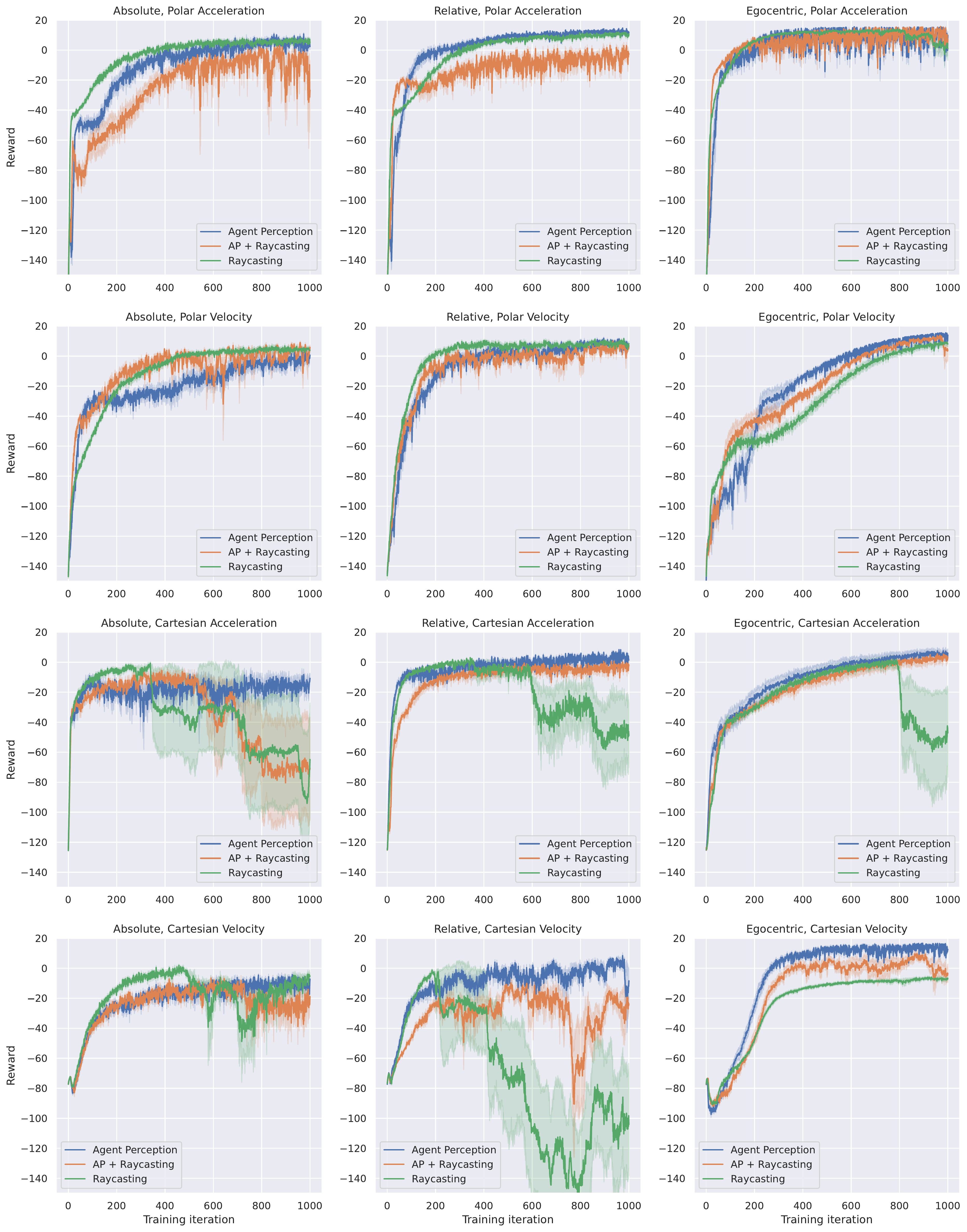}
    \caption{Circle, full range of rewards}
    \label{fig:app-circle-full}
\end{figure*}

\begin{figure*}
    \centering
    \includegraphics[width=\linewidth]{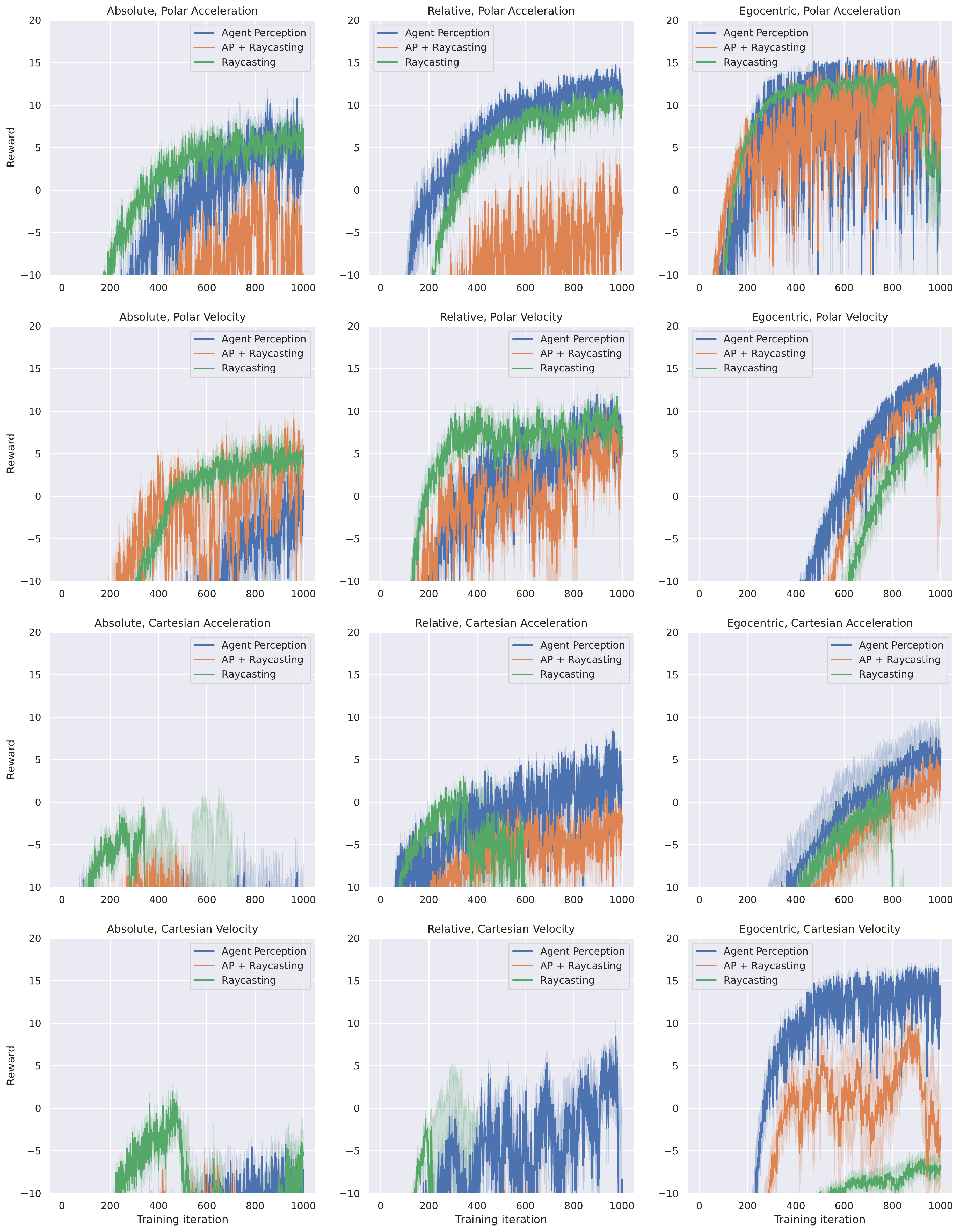}
    \caption{Circle, high rewards}
    \label{fig:app-circle-full}
\end{figure*}

\begin{figure*}
    \centering
    \includegraphics[width=\linewidth]{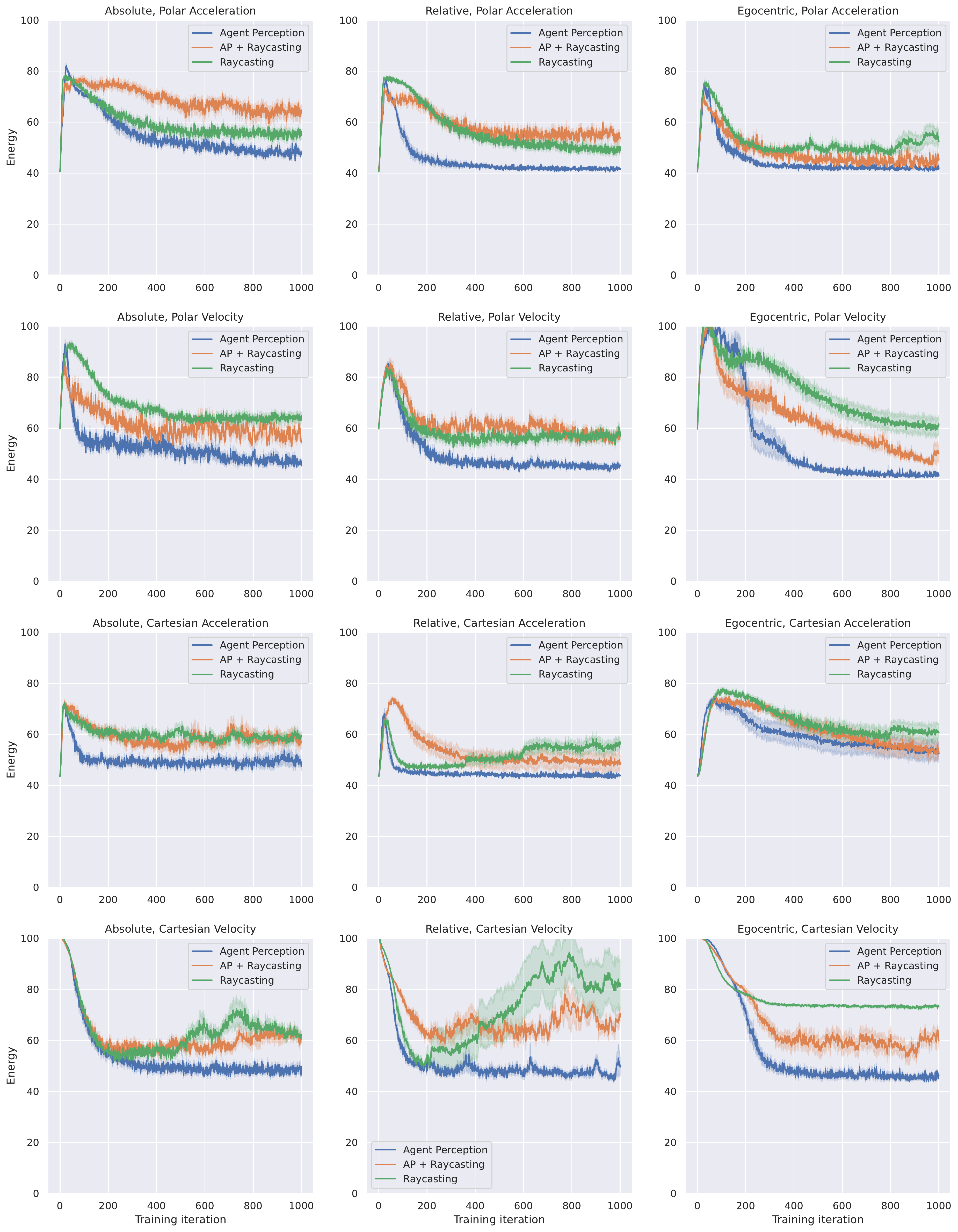}
    \caption{Circle, energy}
    \label{fig:app-circle-energy}
\end{figure*}

\begin{figure*}
    \centering
    \includegraphics[width=\linewidth]{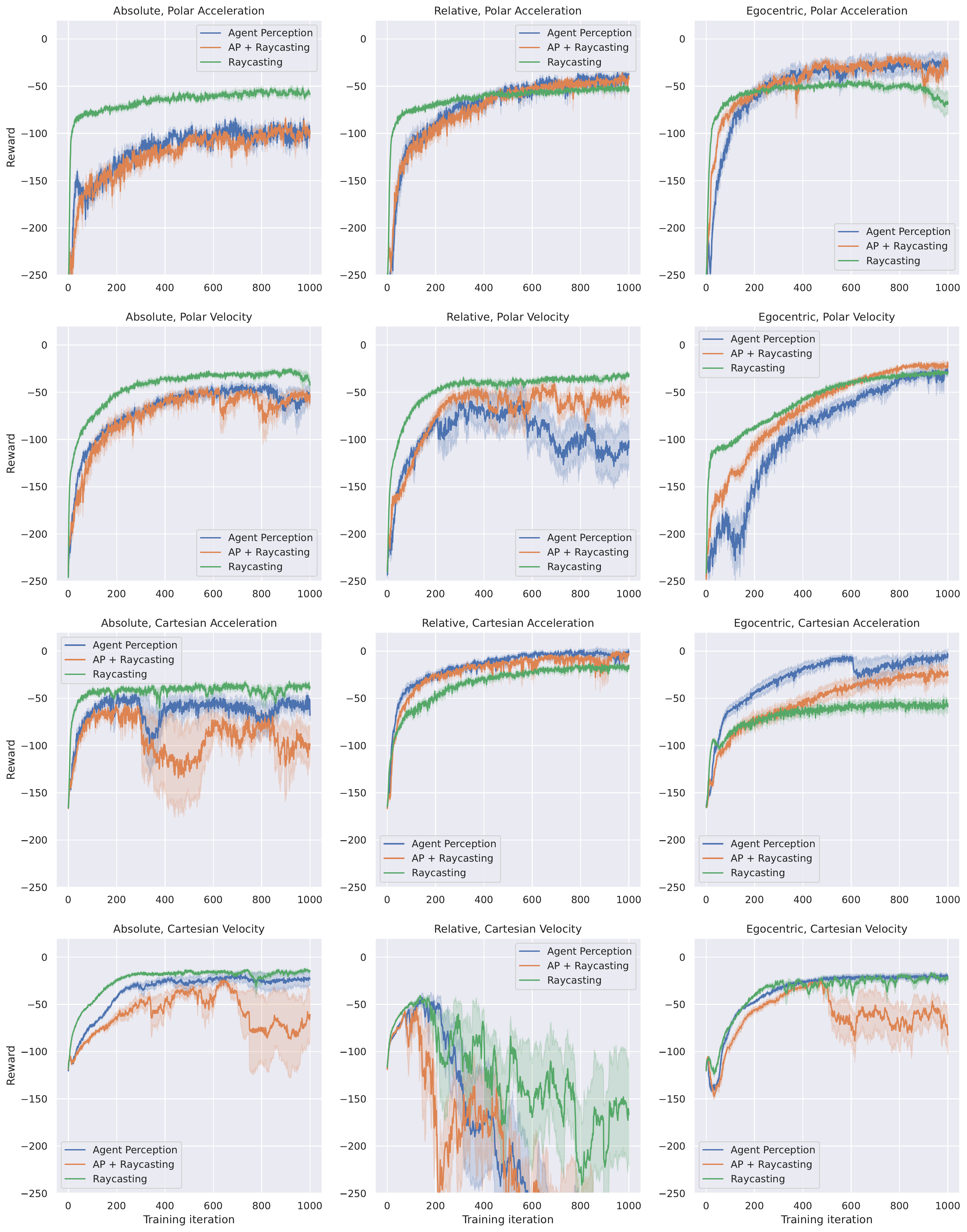}
    \caption{Crossway, full range of rewards}
    \label{fig:app-cross-full}
\end{figure*}

\begin{figure*}
    \centering
    \includegraphics[width=\linewidth]{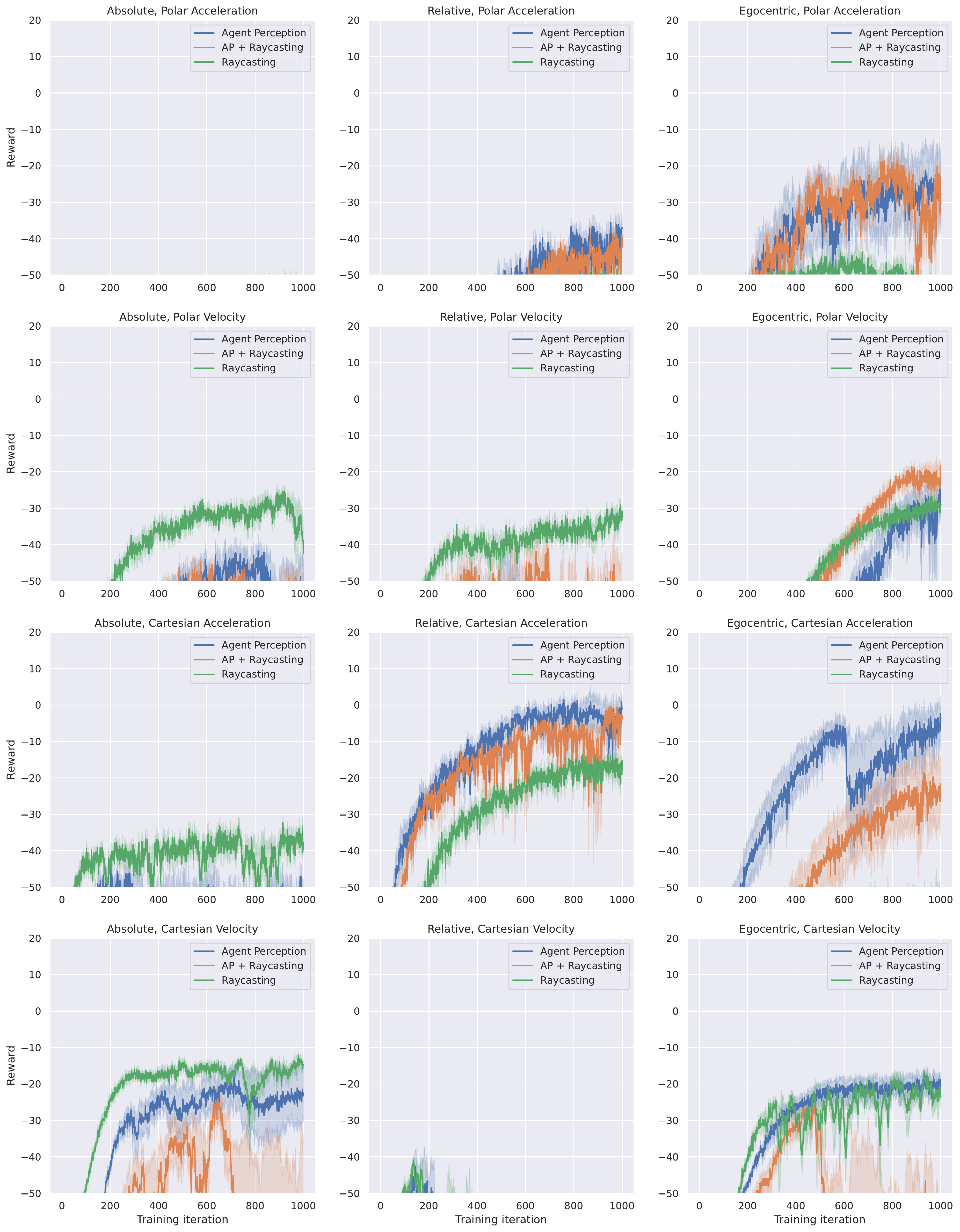}
    \caption{Crossway, high rewards}
    \label{fig:app-cross-full}
\end{figure*}

\begin{figure*}
    \centering
    \includegraphics[width=\linewidth]{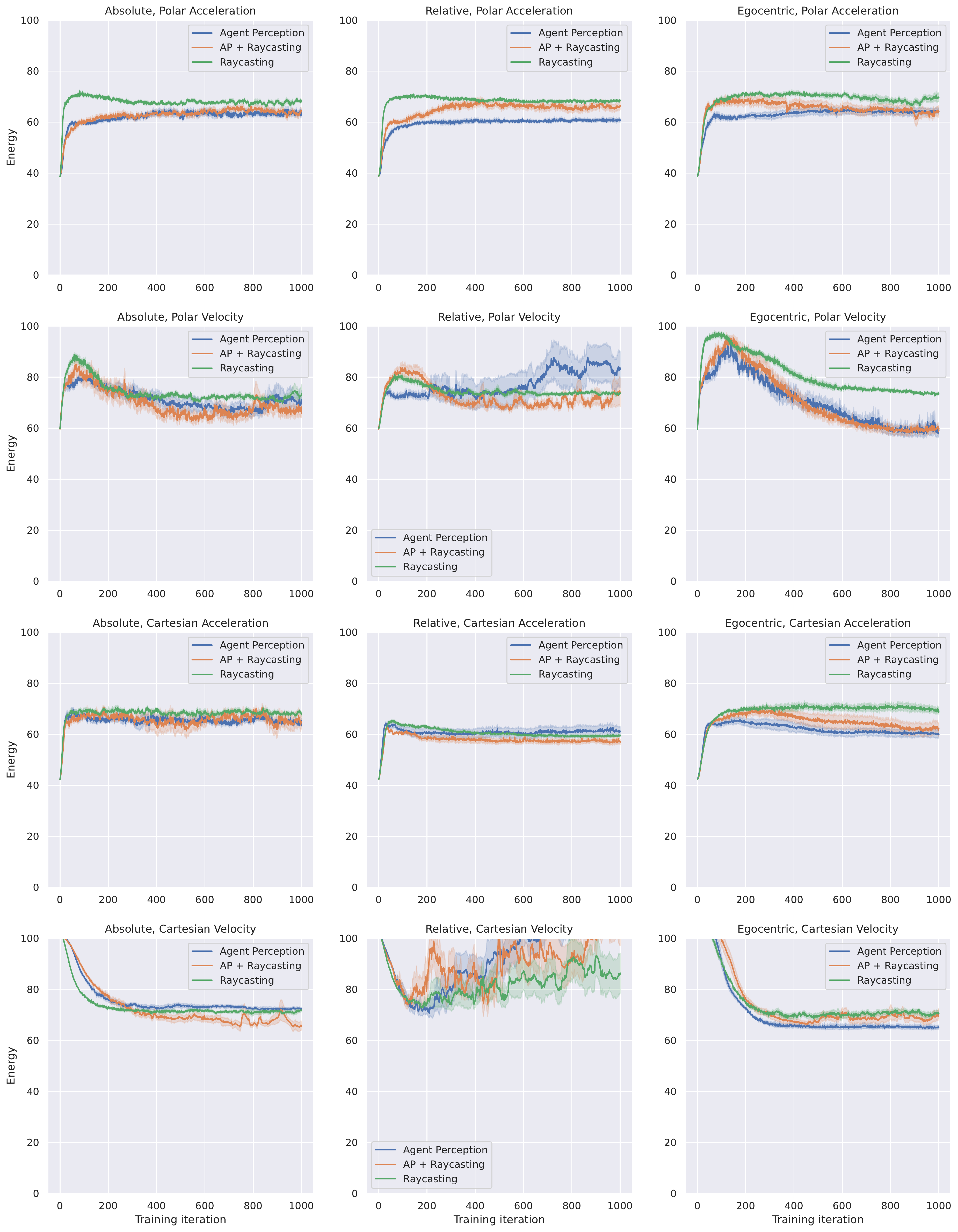}
    \caption{Crossway, energy}
    \label{fig:app-cross-energy}
\end{figure*}

\begin{figure*}
    \centering
    \includegraphics[width=\linewidth]{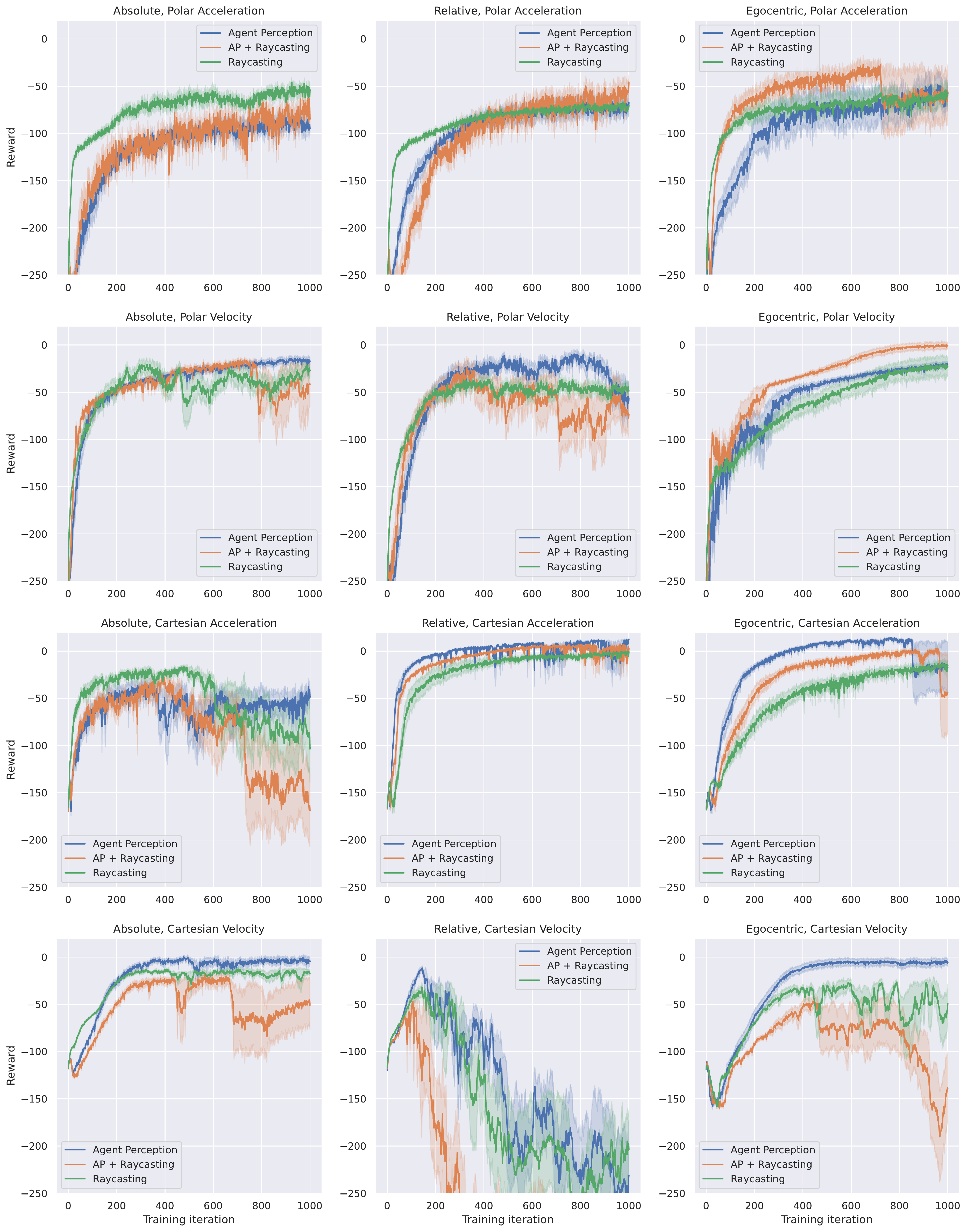}
    \caption{Corridor, full range of rewards}
    \label{fig:app-corridor-full}
\end{figure*}

\begin{figure*}
    \centering
    \includegraphics[width=\linewidth]{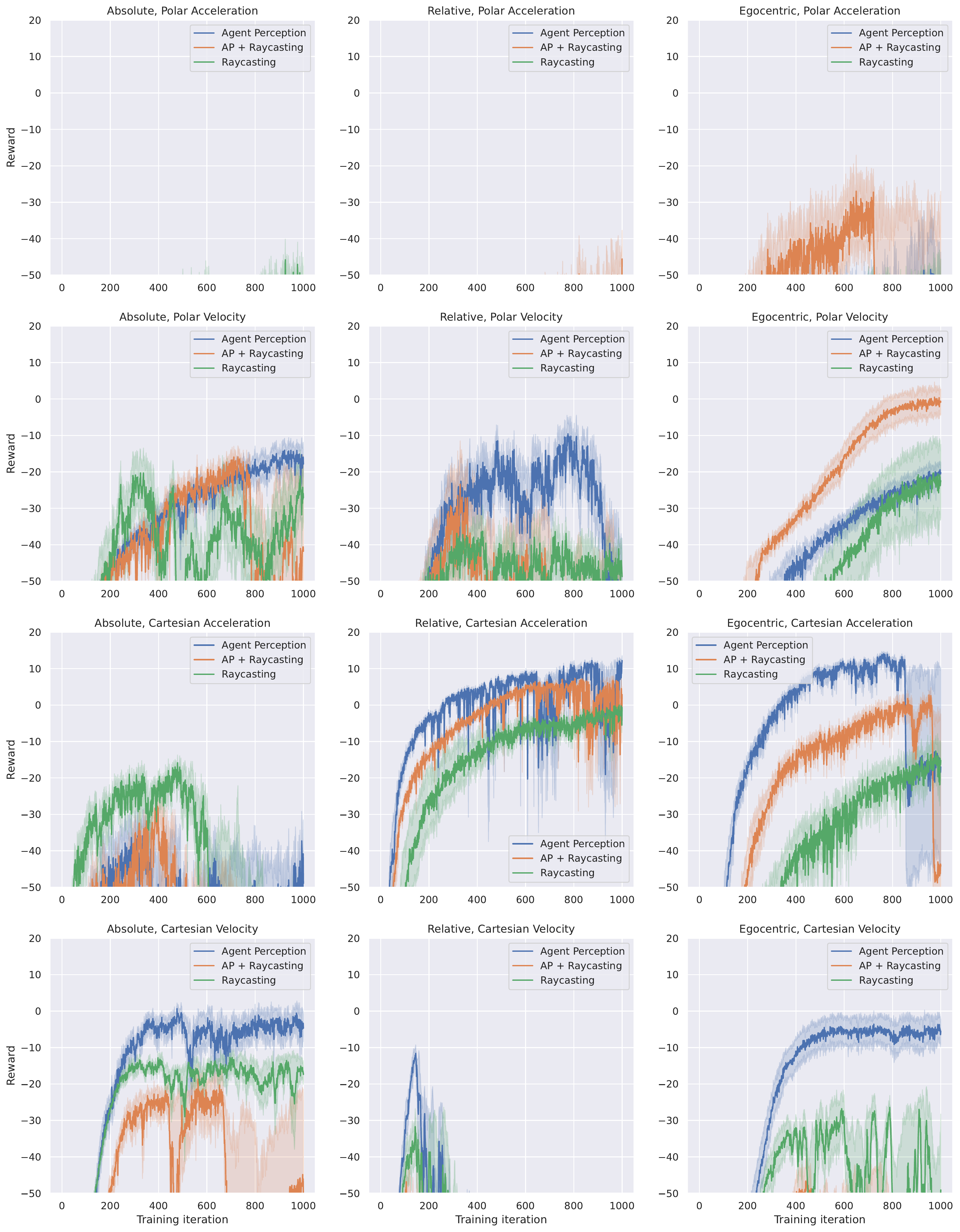}
    \caption{Corridor, high rewards}
    \label{fig:app-corridor-full}
\end{figure*}

\begin{figure*}
    \centering
    \includegraphics[width=\linewidth]{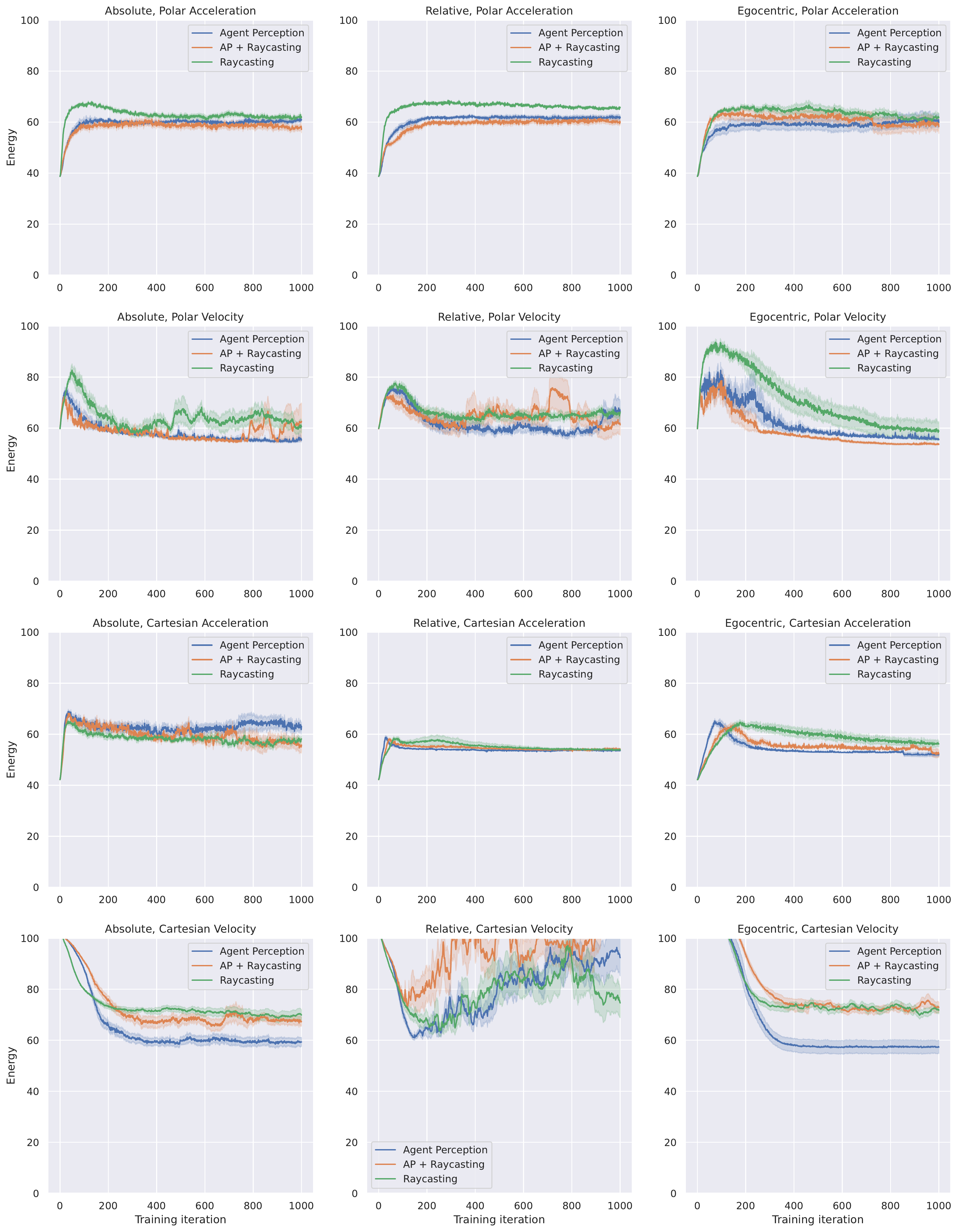}
    \caption{Corridor, energy}
    \label{fig:app-corridor-energy}
\end{figure*}

\begin{figure*}
    \centering
    \includegraphics[width=\linewidth]{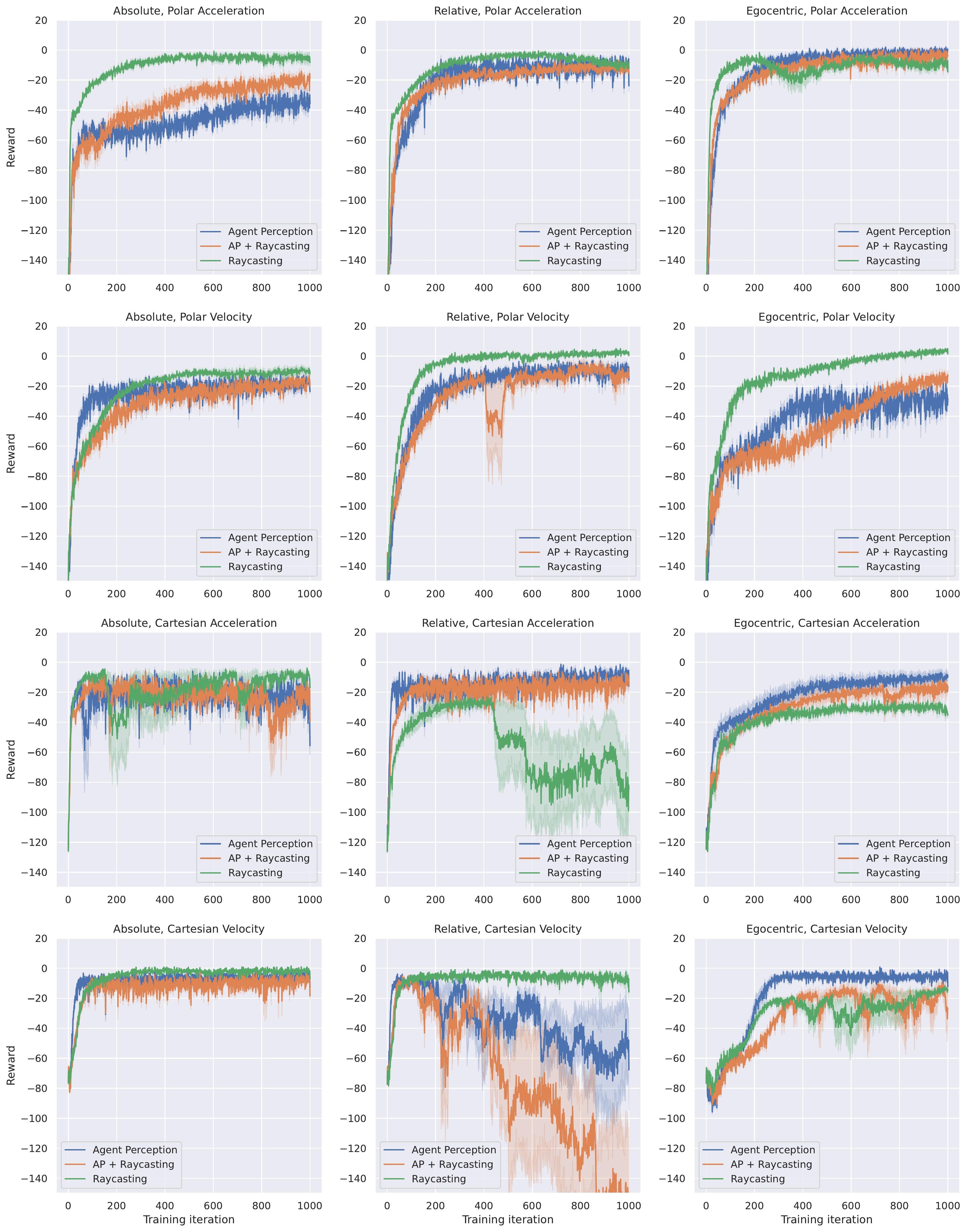}
    \caption{Random, full range of rewards}
    \label{fig:app-random-full}
\end{figure*}

\begin{figure*}
    \centering
    \includegraphics[width=\linewidth]{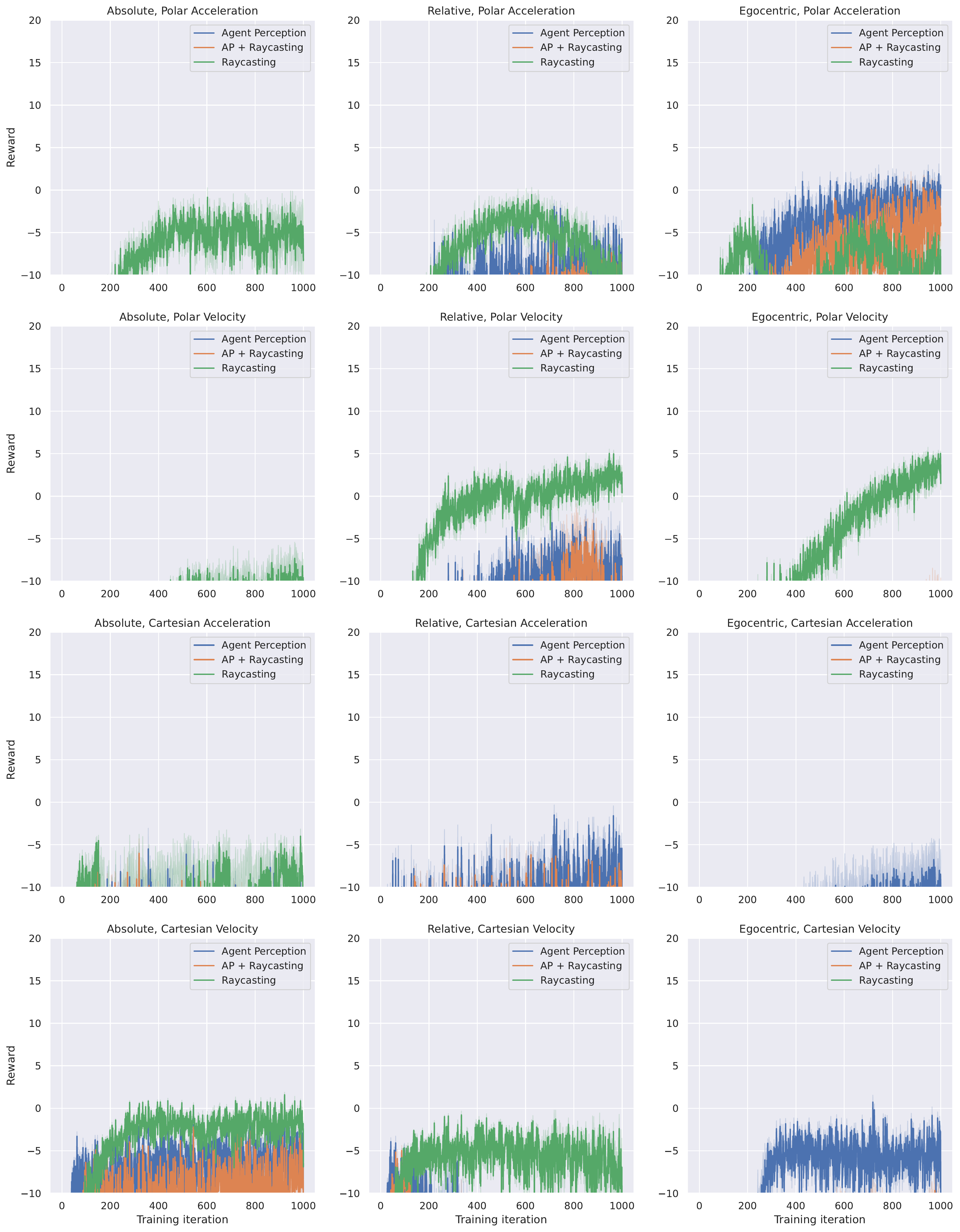}
    \caption{Random, high rewards}
    \label{fig:app-random-full}
\end{figure*}

\begin{figure*}
    \centering
    \includegraphics[width=\linewidth]{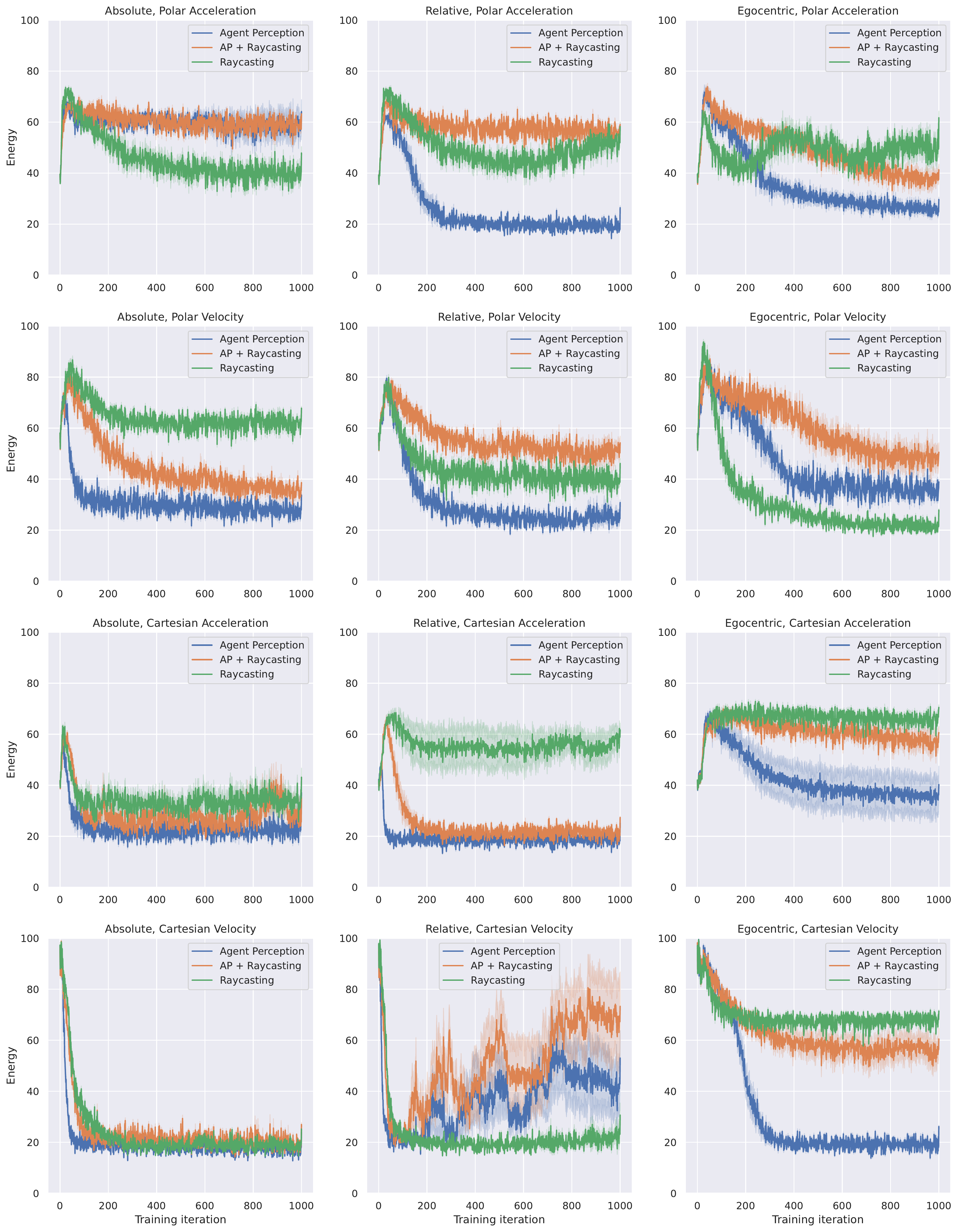}
    \caption{Random, energy}
    \label{fig:app-random-energy}
\end{figure*}

\end{document}